%% file: main.tex
\documentclass[10pt,twocolumn,letterpaper]{article}

\usepackage[pagenumbers]{cvpr} 

\definecolor{cvprblue}{rgb}{0.21,0.49,0.74}
\usepackage[pagebackref,breaklinks,colorlinks,allcolors=cvprblue]{hyperref}

\def\confName{CVPR}
\def\confYear{2025}

\newcommand{\mypara}[1]{\noindent\textbf{#1.}~}

\usepackage[ruled, lined, linesnumbered, commentsnumbered, longend]{algorithm2e}
\usepackage{soul}
\usepackage{multirow}
\usepackage{makecell}
\usepackage{booktabs}
\usepackage{graphicx}
\usepackage{caption}
\usepackage[accsupp]{axessibility}  
\usepackage{cuted}

\title{RigGS: Rigging of 3D Gaussians for Modeling Articulated Objects in Videos}

\author{Yuxin Yao\textsuperscript{1} \quad\quad Zhi Deng\textsuperscript{2} \quad\quad Junhui Hou\textsuperscript{1}\thanks{Corresponding author. This work was supported in part by the NSFC Excellent Young Scientists Fund 62422118, and in part by the Hong Kong RGC under Grants 11219324, 11202320, and 11219422.}\\
	\textsuperscript{1}City University of Hong Kong \quad \textsuperscript{2}University of Science and Technology of China\\
	{\tt\small yuxinyao@cityu.edu.hk} \quad  {\tt\small zhideng@mail.ustc.edu.cn}\quad  {\tt\small jh.hou@cityu.edu.hk} \\ [0.1cm] \url{https://yaoyx689.github.io/RigGS.html}
}

\begin{document}

\maketitle 
\input{figures_scripts/teaser}

\begin{abstract}
This paper considers the problem of modeling articulated objects captured in 2D videos to enable novel view synthesis, while also being easily editable, drivable, and re-posable. To tackle this challenging problem, we propose RigGS, a new paradigm that leverages 3D Gaussian representation and skeleton-based motion representation to model dynamic objects \textit{without} utilizing additional template priors. Specifically, we first propose skeleton-aware node-controlled deformation, which deforms a canonical 3D Gaussian representation over time to initialize the modeling process, producing candidate skeleton nodes that are further simplified into a sparse 3D skeleton according to their motion and semantic information. Subsequently, based on the resulting skeleton, we design learnable skin deformations and pose-dependent detailed deformations, thereby easily deforming the 3D Gaussian representation to generate new actions and render further high-quality images from novel views. Extensive experiments demonstrate that our method can generate realistic new actions easily for objects and achieve high-quality rendering.
\end{abstract}

\input{sections/introduction}
\input{sections/related}

\input{sections/method}

\input{sections/results}
\input{sections/conclusion}

{
    \small
    \bibliographystyle{ieeenat_fullname}
    \bibliography{main}
}

\input{sections/appendix}
\end{document}

%% file: figures_scripts/teaser.tex
\begin{strip}
    \vspace{-6em}
    \centering
    \includegraphics[width=1.0\textwidth]{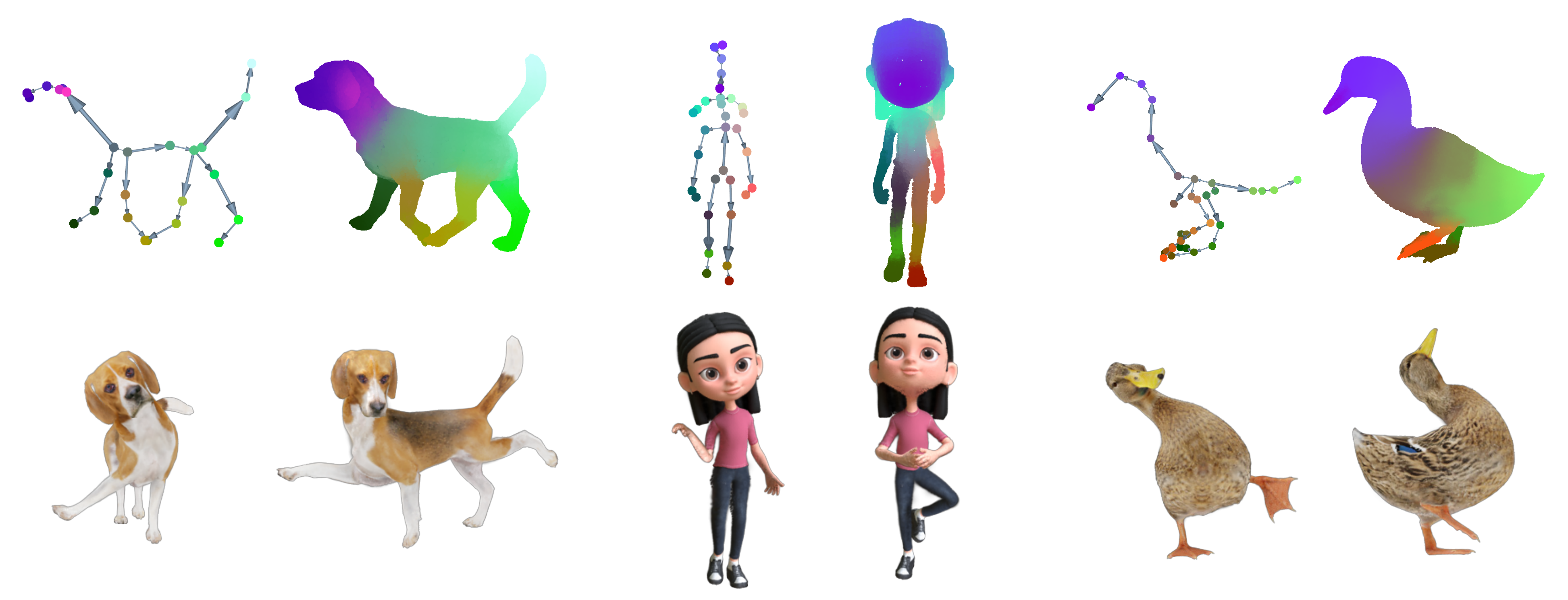}
    \vspace{-2em}
    \captionof{figure}{RigGS is a new and effective paradigm for automatically modeling articulated objects from 2D videos \textit{without} any template prior. RigGS allows for easy editing and interpolation of object motion while supporting high-quality real-time rendering for these creative poses. We visualize the constructed skeleton (top-left), skinning weights (top-right), and edited new poses (bottom) for each object.  
    }
    \label{fig:teaser} 
\end{strip}

%% file: sections/introduction.tex
\section{Introduction}

Rigging involves extracting an animatable skeleton and binding it to a deformable object, which is crucial in movies, games, and AR/VR. The skeleton, composed of interconnected bones and control joints, offers a structured and semantically rich representation of motion. This process is essential for realistic animation, allowing for natural movement and deformation. It also simplifies tasks such as pose editing, motion interpolation, motion transfer, and dynamic animation creation.

Creating semantically plausible rigs automatically is challenging. 
Some methods use universal skeleton models from extensive datasets, like SMPL~\cite{loper15smpl} for the human body, MANO~\cite{romero2017mano} for the human hand, and SMAL~\cite{zuffi2017smal} for animals. 
These models are essential for object reconstruction, pose estimation, animation, etc.
However, many dynamic objects, such as the human body with accessories, hands with personalized gloves, diverse animal species, deformable toys, or robots, cannot be standardized into a universal skeleton. Therefore,
it is highly desired to develop a \textit{template-free} model that can create a rig for any object with a skeletal structure.

Some methods extract skeletons from a 3D representation of the object~\cite{xu2019predicting, xu2020rignet}. However, they rely on artist-designed skeletons as supervision and can only handle symmetric objects. Optimization-based methods~\cite{lin2021point2skeleton,dou2022coverage,wang2024coverage} can extract the skeleton from any 3D model, but the skeletons obtained are often very dense. 
Other approaches combine motion information to extract motion-aware skeletons from point cloud sequences~\cite{xu2022morig,bae2022neural}. However, due to limited 3D data availability, these methods often lack practicality. With the rise of neural rendering, some methods~\cite{wu2023magicpony,yang2023reconstructing} utilize 2D images or videos to obtain rigs for objects. However, they often require a predefined skeleton structure and only optimize joint positions. Other methods~\cite{uzolas2024template,kuai2023camm} do not need a predefined skeleton tree but rely on existing techniques to extract skeletons from 3D models, making them dependent on the quality of the reconstructed geometry and skeleton extraction.

To address these challenges, we propose RigGS, a \textit{template-free} automated rigging model that can extract 3D skeletons from monocular videos of moving objects and bind them to drive deformation. Firstly, we utilize a canonical 3D Gaussian shape representation that is deformed by a skeleton-aware node-controlled deformation field over time to initialize the modeling process, producing a set of skeleton-aware nodes. We then simplify the dense nodes using geometric, semantic, and motion information to achieve the final sparse 3D skeleton. Finally, we bind the canonical 3D Gaussians to the 3D skeleton to create a skeleton-driven dynamic model. Once trained, RigGS enables editing, motion interpolation, motion transfer, and animation of dynamic objects while supporting high-quality real-time rendering.
Extensive experiments indicate that our RigGS can extract reasonable 3D skeletons, achieves rendering accuracy comparable to state-of-the-art methods, and allows for the flexible generation of new motions for the reconstructed dynamic objects.
  
In summary, we propose a novel \textit{template-free} paradigm that can synthesize dynamic objects captured in  2D videos from novel views and facilitate editing to create new actions. The technical novelty of our approach lies in: 
\begin{itemize}

\item we introduce a deformation field based on skeleton-aware nodes, combined with 3D Gaussians as the canonical shape representation, achieving simultaneous reconstruction of dynamic objects and obtaining candidate skeleton points;
\item we propose a heuristic 3D skeleton construction algorithm that considers geometric, semantic, and motion information; and
\item we develop a skeleton-driven dynamic model with learnable skinning weights to bind the skeleton with the 3D Gaussian representation and a pose-dependent detailed deformation, allowing for flexible generation of new motions.
\end{itemize}

%% file: sections/related.tex
\section{Related work}

\mypara{Geometric Representation and Rendering}
Traditional 3D modeling utilizes textured meshes as a typical representation. With the emergence of neural implicit representations, many methods are employing them to model 3D objects or scenes~\cite{mildenhall2020nerf,wang2021neus}. Recently, 3D Gaussian Splatting (3DGS)~\cite{kerbl20233dgs} has gained attention for its efficient and high-quality rendering capabilities, along with its ease of editing, gradually becoming an important representation in 3D vision. Numerous approaches~\cite{wu20244dgs,huang2024sc,yang2024deformable,you2024decoupling,liu2024modgs,you2024nvs} have also emerged to address the task of novel view synthesis for dynamic objects or scenes captured by videos.

\vspace{0.5em}
\mypara{Prior-dependent Dynamic Modeling}
Considering the challenges in skeleton extraction, various methods have leveraged category priors to establish specific parameterized models. For example, SMPL~\cite{loper15smpl} is designed to represent human bodies, MANO~\cite{romero2017mano} focuses on hand modeling, SMAL~\cite{zuffi2017smal} extends to handle quadrupeds, and so on.
These parameterized models have played important roles in tasks such as reconstruction, tracking, and animation within their domains~\cite{jiang2022selfrecon,lei2024gart,moon2024exavatar,yang2024omnimotiongpt}.
However, their representational capabilities are confined when dealing with shapes beyond their intended distributions, such as humans in intricate attire or animals from diverse categories.
To enable broader applications, some methods do not utilize these parameterized models. 
CASA~\cite{wu2022casa} establishes a database of animals, enabling category-agnostic skeletal animal reconstruction from monocular videos by retrieving and refining similar 3D models and skeletons from the database. 
Some methods take a pre-established skeletal tree structure as input, optimizing the positions of skeletal joints to model objects~\cite{yao2022lassie,wu2023magicpony,yang2023rac}. 
However, the pre-definition of the skeletal tree structure hampers the automation of skeleton creation.

\vspace{0.5em}
\mypara{Neural Bones for Dynamic Objects}
To establish a deformation field for arbitrary objects, BANMo~\cite{yang2022banmo} introduces a representation using neural bones. It no longer relies on the traditional skeletal tree structure; instead, it employs learnable bones, involving spatial positions and transformations, to represent the deformation.
These neural bones can be distributed on the surface, inside, or even outside of the object.
Similar to some control point-based deformation fields~\cite{bozic2021ndg,huang2024sc,yao2024dynosurf}, these bones do not carry semantic information. Following BANMo, BAGS~\cite{zhang2024bags} adopt diffusion prior and 3DGS to construct animatable model from a single casual video. DreaMo~\cite{tu2023dreamo} further establishes the skeleton based on the learned bones. It derives the connections between bones through clustering, utilizing the reconstructed mesh and skinning weights.
Compared with the traditional skeleton structure, it is easier to construct, but more difficult to create reasonable new actions.

\vspace{0.5em}
\mypara{Template-free Skeleton for Articulated Objects}
Establishing a skeleton tree with a template-free algorithm for dynamic objects, thereby offering enhanced semantic meaning and editability is extremely challenging. 
Some methods utilize complete 3D representations, such as meshes, to achieve this~\cite{xu2019predicting,xu2020rignet}. 
However, they can only handle meshes that have a symmetrical structure.
Some methods focus on handling objects with articulated rigid structures~\cite{liu2023reart,song2024reacto}, such as laptops, glasses, or drawers. However, they cannot handle other types of dynamic objects, such as animals. 
Some methods take 2D images as input to build animatable models~\cite{kuai2023camm,uzolas2024template}. They first reconstructed the dynamic representation and extracted the 3D mesh of the template shape, using the skeleton extraction method of the 3D model mentioned above to obtain the skeleton. Even though the skeleton will be refined in subsequent optimizations, they still rely on the quality of the 3D skeleton extraction and reconstruction of the template shape. Recently, SK-GS~\cite{wan2024skgs} utilizes 3D Gaussian Splatting and super-points to reconstruct dynamic objects and discovers the underlying skeleton model by treating super-points as rigid parts.

%% file: sections/method.tex
\input{figures_scripts/pipeline}

\section{Proposed Method}

\textbf{Overview.} Given a monocular video capturing continuous actions of an object, denoted as $\mathcal{I}=\{\mathbf{I}_t\}$ with $\mathbf{I}_t$ being the frame at time $t\in[0,~1]$,
we aim to model the animatable dynamic object such that it can be rendered from novel viewpoints,
 and \textit{more importantly, the object can be easily edited to create new actions}.  
To tackle this challenging problem, we propose an automated 3D skeleton construction method along with a skeleton-driven deformation model.
As illustrated in Fig.~\ref{fig:pipeline}, we initially model  
the dynamic object through 3D Gaussian splatting in conjunction with a skeleton-aware node-controlled deformation model, resulting in a 3D Gaussian-based canonical representation and nodes imbued with skeleton semantics. These skeleton-aware nodes allow for the generation of a dense and redundant skeleton model, which is further simplified and enhanced based on geometric, motion, and symmetry considerations. Finally, we introduce a skeleton-driven deformation model that finely fits the input video and generates new actions.

\subsection{Initialization}
\label{sec:initial_reconstruction}
At this stage, we initialize the 4D reconstruction of the dynamic scene captured in $\mathcal{I}$ using a canonical 3D Gaussian representation that is deformed over time by skeleton-aware node-controlled deformation.   

\vspace{0.5em}
\mypara{Canonical 3D Gaussian Representation}
The 3D Gaussian representation \cite{kerbl20233dgs} is a collection of 
attributed 3D Gaussians with each Gaussian $G_i$ containing a center position ${\mu}_i$, covariance matrix $\boldsymbol{\Sigma}_i$, opacity ${\sigma}_i$ and spherical harmonic coefficient $sh_i$.
The covariance matrix $\boldsymbol{\Sigma}_i$ can be decomposed as $\boldsymbol{\Sigma}_i=\mathbf{R}_i\mathbf{S}_i\mathbf{S}_i^T\mathbf{R}_i^T$ for optimization, where $\mathbf{R}_i$ is a rotation matrix represented by a quaternion $\mathbf{q}_i$, and $\mathbf{S}_i$ is a scaling matrix denoted by a 3D vector $\mathbf{s}_i$.
Rendering an image from a specific viewpoint $v_i$ involves projecting the 3D Gaussians onto a 2D plane, resulting in 2D Gaussians with projected means $\hat{{\mu}}_i$ and covariances $\hat{\boldsymbol{\Sigma}}_i$.
The color $\mathcal{C}(u)$ of the image pixel $u$ can be calculated by 
\begin{equation}
\mathcal{C}(u) = \sum_{i\in N} T_i \alpha_i \mathcal{SH}(sh_i, v_i), \text{~~where~} T_i=\prod_{j=1}^{i-1}(1-\alpha_j).
\end{equation}
Here $\mathcal{SH}(\cdot, \cdot)$ is the spherical harmonic function, and $\alpha_i$ can be calculated by\vspace{-0.3cm}
\begin{equation}
\alpha_i = \sigma_i \exp({-\frac{1}{2}(u-\hat{{\mu}}_i)^T\hat{\boldsymbol{\Sigma}}_i(u-\hat{{\mu}}_i)}).
\end{equation}
Therefore, a 3D scene is parameterized as $\mathcal{G}=\{G_i: \mu_i, \mathbf{q}_i, \mathbf{s}_i, \sigma_i, sh_i\}$, which will be adjusted adaptively during the optimization process.

\vspace{0.5em}
\mypara{Skeleton-aware Node-controlled  Deformation} 
This module facilitates the temporal deformation of the canonical 3D Gaussian representation to model the dynamic scene. Specifically, the deformation of each 3D Gaussian at time $t$ is achieved as follows:
\begin{equation}
\label{eq:deform-mu}
\mu_i^t = \sum_{\mathbf{c}\in\mathcal{N}(\mu_i)} w_{\mu_i, \mathbf{c}}(\tilde{\mathbf{R}}_{\mathbf{c}}^t(\mu_i-\mathbf{c}) + \mathbf{c} + \tilde{\mathbf{t}}_{\mathbf{c}}^t),
\end{equation}
where $\mathbf{C} = \{\mathbf{c}\}$ denotes the set of skeleton-aware nodes; 
$\tilde{\mathbf{R}}_{\mathbf{c}}^t\in \mathbb{R}^{3\times 3}$ and $\tilde{\mathbf{t}}_{\mathbf{c}}^t\in\mathbb{R}^{3}$ are the rotation matrix and translation vector of node $\mathbf{c}$ at time $t$; $\mathcal{N}(\mu_i)$ is the set of the $k$ nearest points to $\mu_i$ in $\mathbf{C}$; 
the weight $w_{\mu_i, \mathbf{c}}$ is defined as\vspace{-0.2cm} 
\begin{equation}
w_{\mu_i, \mathbf{c}} = \frac{w_{\mu_i, \mathbf{c}}'}{\sum_{\mathbf{c}\in\mathcal{N}(\mu_i)}w_{\mu_i, \mathbf{c}}'}, 
\end{equation}
where 
\begin{equation}
w_{\mu_i, \mathbf{c}}' = \exp \left(-\frac{\|\mu_i - \mathbf{c}\|_2^2}{2o_{\mathbf{c}}^2}\right).
\end{equation}
Here $o_{\mathbf{c}}$ is a learnable radius. 
The deformed node can be computed as  
\begin{equation}
\mathbf{c}^t = \mathbf{c} + \tilde{\mathbf{t}}_{\mathbf{c}}^t. 
\end{equation}
In cases where the 3D Gaussian exhibits anisotropy, its rotation at time $t$ can be defined as\vspace{-0.2cm}
\begin{equation}
\label{eq:deform-q}
\mathbf{q}_i^t = (\sum_{\mathbf{c}\in\mathcal{N}(\mu_i)} w_{\mu_i, \mathbf{c}}\mathbf{r}_{\mathbf{c}}^t)\otimes \mathbf{q}_i,\vspace{-0.25cm}
\end{equation}
where $\mathbf{r}_{\mathbf{c}}^t\in\mathbb{R}^4$ is the quaternion representation of predicted rotation; $\otimes$ is the production of quaternions; and $\tilde{\mathbf{R}}_{\mathbf{c}}^t,\tilde{\mathbf{t}}_{\mathbf{c}}^t$ and $\mathbf{r}_{\mathbf{c}}^t$ will be learnable via an MLP parameterized with $\Theta$, denoted as $F_{\Theta}(\gamma_{\mathbf{c}}(\mathbf{c}),\gamma_t(t))$ with $\gamma_{\mathbf{c}}$ and $\gamma_t$ being the positional encoding.

\vspace{0.5em}
\mypara{Loss Function for Optimizing the Initialization Stage}
By capitalizing on the alignment of the skeleton with an object's central axis, we introduce 2D skeleton projection constraints to refine the skeleton-aware nodes.
Initially, we derive a set of 2D skeleton points, denoted as $\{\mathbf{p}_j^t\}$, from a reference silhouette using a skeletonize/morphological thinning algorithm~\cite{zhang1984thinning}. 
For deformed nodes $\{\mathbf{c}^t\}$ at time $t$, we project them onto the 2D plane based on the camera viewpoint $v_t$ and then calculate the projection loss as:
\begin{equation}
\label{eq:2dproj-loss}
L_{\texttt{proj}}^t = \mathcal{CD}_{\ell_1}(\texttt{Proj}(\mathbf{c}^t, v_t), \{\mathbf{p}_j^t\}) ,
\end{equation}
where $\texttt{Proj}(\cdot,\cdot)$ is a projection operator, and $\mathcal{CD}_{\ell_1}(\cdot,~\cdot)$ is the $\ell_1$-norm-based Chamfer distance~\cite{fan2017point}. Notably, unlike approaches that directly extract the central axis from the reconstructed 3D model~\cite{uzolas2024template}, our method is simpler and not reliant on the 3D reconstruction quality.

Moreover, we compute the rendering loss $L_{\texttt{render}}^t$ between the rendering image $\hat{\mathbf{I}}_t$ at time $t$ and input image $\mathbf{I}_t$ using a combination of $\ell_1$ loss and D-SSIM loss~\cite{kerbl20233d}. Inspired by SC-GS~\cite{huang2024sc}, we also introduce the ARAP loss~\cite{sorkine2007rigid} $L_{\texttt{arap}}^t$ as a regularization term to maintain local rigidity during deformation.

In summary, the overall loss at time $t$ is $L^t_{\texttt{init}} = L_{\texttt{render}}^t + w_{\texttt{arap}} L_{\texttt{arap}}^t + w_{\texttt{proj}}L_{\texttt{proj}}^t$, where $w_{\texttt{arap}}$ and $w_{\texttt{proj}}$ are weights to balance the these terms. 

\subsection{Coarse-to-Fine 3D Skeleton Construction}
\label{sec:skeleton_construction}
Referring to the initial dynamic reconstruction acquired in Sec.~\ref{sec:initial_reconstruction}, we begin by choosing a fresh canonical shape, followed by introducing a heuristic algorithm for deriving a sparse skeleton from the skeleton-aware nodes linked to the chosen canonical shape.

\vspace{0.5em}
\mypara{Selection of Canonical Shape} 
Since the canonical shape derived in Sec.~\ref{sec:initial_reconstruction}  doesn't correspond to any specific frame and hence lacks physical meaning (see Fig.~\ref{fig:pipeline}), 
we replace it with a deformed shape at a chosen time representing the mean shape. Benefiting from the initialized reconstruction, we select the canonical shape based on pre-computed motion trajectories of skeleton-aware nodes. 
Specifically, for each node $\mathbf{c}$, we compute the mean of its trajectory, that is 
$
\overline{\mathbf{c}} = \sum_{\mathbf{I}_t\in \mathcal{I}}\mathbf{c}^t / |\mathcal{I}|,
$ 
and select the time for the canonical shape by 
\[t^*=\mathop{\arg\min}_{t}\sum_{\mathbf{c}}\|\mathbf{c}^t-\overline{\mathbf{c}}\|_2.\] 
In the following, the deformed Gaussians $\{G_i^{t^*}\}$ and node $\mathbf{c}^{t^*}$ at time $t^*$ will serve as the new canonical shape and skeleton candidate points.

\input{figures_scripts/skeleton_construction}

\vspace{0.5em}
\mypara{Dense Skeleton Construction}
To remove redundant nodes, we initially employ farthest point sampling (FPS) on $\{\mathbf{c}^{t^*}\}$ to derive a subset of nodes that are uniformly distributed $\{\mathbf{c}_s^{t^*}\}_{s\in\mathcal{S}}$ with $\mathcal{S}$ being the index set. To establish the edge set to form a tree structure, we first construct an undirected fully connected graph with edge weights 
$\{\beta_{ij}\}_{i,j\in\mathcal{S}}$ for $\{\mathbf{c}_s^{t^*}\}$. We define
\[
\beta_{ij}=\sum_{t}d_{ij}^{t}/|\mathcal{I}|,~~\text{where}~~d_{ij}^t=\|\mathbf{c}_i^t - \mathbf{c}_j^t\|,
\]
considering both motion and positional information. 
We then use Prim’s algorithm to obtain the minimum spanning tree. 
This tree contains three node types: junctions ($>2$ neighbors), endpoints ($1$ neighbor), and connection points ($2$ neighbors).
Due to noise in $\{\mathbf{c}_s^{t^*}\}$, unnecessary bifurcations may occur. Therefore, we remove redundant branches from junctions to endpoints and merge two closely located junctions.  
The dense skeleton tree is then constructed, denoted as $\mathcal{T}_d=\{\mathcal{J}_d, \mathcal{E}_d\}$, where $\mathcal{J}_d=\{\mathbf{J}_i\}$ denotes the node positions, and $\mathcal{E}_d$ represents the connection betweens the nodes. 

\vspace{0.5em}
\mypara{Skeleton Simplification}
For a more concise deformation representation,  we devise a heuristic algorithm to obtain a sparse skeleton. We start with an empty sparse joint set, denoted as $\mathcal{J}$, and progressively incorporate probable joint nodes from $\mathcal{J}_d$. Endpoints and junctions are typically more meaningful,
so we first add them into $\mathcal{J}$. 
We then select points with a high probability of being geometric turning points in $\mathcal{J}_d$, forming a potential joint point set $\mathcal{H}$. Using semantic labels from DINOv2 features~\cite{oquab2023dinov2}, we ensure semantic symmetry by adding or removing points from $\mathcal{H}$. Finally, we incorporate $\mathcal{H}$ into $\mathcal{J}$ to form the final joint set.
Using breadth-first search, we find the nearest endpoint for each junction, record the path length, and designate the junction with the longest path as the root joint $\mathbf{J}_r$. 
Starting from the root joint, we define the joint closer to it as the parent and the other as the child for each edge.
The sparse skeleton tree is denoted as $\mathcal{T}=\{\mathcal{J}, \mathcal{A}\}$, where $\mathcal{A}=\{A_j\}_{\mathbf{J}_j\in{\mathcal{J} \setminus \{\mathbf{J}_r}\}}$ contains all parent indices. 
Fig.~\ref{fig:skeleton_construct} illustrates the skeleton construction process. More details can be found in the \textit{Supplementary Material}.

\subsection{Skeleton-driven Dynamic Modeling}
Built upon the initial reconstruction in Sec.~\ref{sec:initial_reconstruction}, the new canonical shape and the skeleton obtained in Sec.~\ref{sec:skeleton_construction}, we establish a skeleton-driven dynamic model including an LBS-based coarse deformation with learnable skinning weights and a pose-dependent detail deformation model at this stage.

\vspace{0.5em}
\mypara{Learnable LBS-based Coarse Deformation}
Denote by $b_{j}$ the edge/bone between joint $\mathbf{J}_j$ and its parent $\mathbf{J}_{A_j}$. Similar to~\cite{wu2023magicpony,uzolas2024template}, 
we define global translation $\hat{\mathbf{t}}^t$ and rotation transformations $\{\hat{\mathbf{R}}_{j}^t\}_{\mathbf{J}_j\in\mathcal{J}}$ at time $t$, representing the global rotation for the root $\mathbf{J}_r$ and the rotation of the child joint $\mathbf{J}_j$ around the parent joint $\mathbf{J}_{A_j}$ for others. Without loss of generality, we define $\mathbf{J}_1$ as the root node. 
For the center of a Gaussian $\mu_i$ in the canonical shape, we can deform it using linear blend skinning (LBS)~\citep{lewis2023pose}:  
\begin{equation}
\label{eq:skeleton-deform}
\hat{\mathbf{\mu}}_i^t = \mathbf{T}_1^t\left(\sum_{j=2}^{|\mathcal{J}|} \hat{\omega}_{i,j} \mathbf{T}_j^t \overline{\mu}_i\right), \vspace{-0.3cm}
\end{equation}
where 
\begin{equation}
\mathbf{T}_j^t=\mathbf{T}_{A_j}^t \hat{\mathbf{T}}_{j}^t 
~~\text{and}~~\hat{\mathbf{T}}_j^t = 
\begin{bmatrix}
\hat{\mathbf{R}}_{j}^t & \mathbf{J}_{A_j} - \hat{\mathbf{R}}_{j}^t \mathbf{J}_{A_j} \\
\mathbf{0} & 1
\end{bmatrix},
\end{equation}
and 
\[\mathbf{T}_1^t=
\begin{bmatrix}
\hat{\mathbf{R}}_1^t & \mathbf{J}_1 - \hat{\mathbf{R}}_1^t \mathbf{J}_{1} + \hat{\mathbf{t}}^t \\
\mathbf{0} & 1 
\end{bmatrix}.
\]
Here $\mathbf{T}_j^t$ is defined recursively by its parent $\mathbf{T}_{A_j}^t$; $\overline{\mu}_i$ is the homogeneous coordinate representation of $\mu_i$;
$\hat{\omega}_{i,j}$ is the learnable skinning weight, which can be calculated by 
\begin{equation}
\label{eq:skinning-weights}
\hat{\omega}_{i,j} = \frac{\tilde{\omega}_{i,j}}{\sum_{j=2}^{|\mathcal{J}|}\tilde{\omega}_{i,j}},
\quad\text{where~} \tilde{\omega}_{i,j} = \eta_{i,j}\exp(-\frac{D^2(\mu_i, b_j)}{2\nu_j^2}).
\end{equation}
Here $D(\mu_i, b_j)$ is the distance between the Gaussian center $\mu_i$ and the bone $b_j$;  $\{\nu_j\}_{j=2}^{|\mathcal{J}|}
$ are the learnable radius parameters; $\{\eta_{i,j}\}_{j=2}^{|\mathcal{J}|}$ are the learnable scaling factors and are learned by an MLP $F_{\Psi}(\gamma_{\eta}(\mu_i))$ parameterized with $\Psi$; $\{\hat{\mathbf{R}}_1^t,...,\hat{\mathbf{R}}_{|\mathcal{J}|}^t\}$ are represented as in quaternion form $\{\hat{\mathbf{r}}_1^t,...,\hat{\mathbf{r}}_{|\mathcal{J}|}^t\}$ , together with translation $\hat{\mathbf{t}}^t$, learned by an MLP $F_{\Phi}(\gamma_t(t))$ parameterized with $\Phi$.  
It can be used to correct inappropriate geometric distance-based weights by incorporating motion information.
When the 3D Gaussian exhibits anisotropy, we approximate its rotation by the rotation part of $\mathbf{T}_1^t\left(\sum_{j=2}^{|\mathcal{J}|} \hat{\omega}_{i,j} \mathbf{T}_j^t \right)$. 

\vspace{0.5em}
\mypara{Pose-dependent Detail Deformation}
Due to the sparsity of the skeleton, the deformation field it represents exhibits local rigidity, limiting its effectiveness in areas with fine details, such as cloth wrinkles undergoing subtle deformations under physical forces. To address this, we incorporate a pose-dependent detail deformation module that learns local details. 
Moreover, making this module pose-related rather than time-related allows us to generate more plausible details when creating new actions. Specifically, we use an MLP parameterized with $\Pi$, denoted as $F_{\Pi}(\mu_i, \{\hat{\mathbf{r}}_1^t,...,\hat{\mathbf{r}}_{|\mathcal{J}|}^t\})$,  to learn the offsets $\delta_{i,t}$ for the center position $\mu_i$ of a Gaussian. 
The final position of the Gaussian center is then computed as $\mu_i^t=\hat{\mu}_i^t + \delta_{i,t}$.  
By combining this with the rendering formula of 3D Gaussian splatting, we can obtain images from novel viewpoints and generate new motions by changing the pose.

\vspace{0.5em}
\mypara{Loss Function} 
Apart from the rendering loss $L_{\texttt{render}}^t$ detailed in Sec.~\ref{sec:initial_reconstruction}, we introduce three additional loss terms, resulting in the overall loss function at time $t$:
\[L^t= L_{\texttt{render}}^t + w_{\tilde{\texttt{proj}}}^t L_{\tilde{\texttt{proj}}}^{t} + w_{\texttt{detail}}^t L_{\texttt{detail}}^t + w_{\texttt{id}} L_{\texttt{id}}^{t}.\]
To ensure the deformed skeleton remains within the deformed shape, we use a skeleton projection loss similar to Eq.~\eqref{eq:2dproj-loss} in Sec.~\ref{sec:initial_reconstruction}. 
Because the skeleton joints are sparser than the skeleton-aware nodes, we sample $\mathbf{Q}_{\mathcal{T}}^t$ along the skeleton bones to compute the Chamfer distance loss:
\begin{equation}
L_{\tilde{\texttt{proj}}}^t = \mathcal{CD}_{\ell_1}(\texttt{Proj}(\mathbf{Q}_{\mathcal{T}}^t, v_t), \{\mathbf{p}_j^t\}).
\end{equation}
Due to the imperfect accuracy of the 2D skeleton (see Fig.~\ref{fig:ablas_projskel}), these inaccurate 2D skeletons can lead to erroneous guidance in deformation. Therefore, we design time-specific weights: $w_{\tilde{\texttt{proj}}}^t = 10^{-3}\cdot\exp(-(L_{\tilde{\texttt{proj}}}^t)^2/2\xi^2)$, where $\xi$ is half the median of $\{L_{\tilde{\texttt{proj}}}^t\}_t$. According to the 3-sigma rule of the Gaussian distribution, 2D skeletons with errors less than 1.5 times the median are considered accurate and will be subjected to stronger constraints. 

To handle significant movements through LBS-based deformation and capture finer details with the other module, we introduce the regularization term
\begin{equation}
L_{\texttt{detail}}^t = \sum_{i}\|\delta_{i,t}\|_2^2 /|\mathcal{G}|,
\end{equation}
aimed at minimizing detailed deformations, where $|\mathcal{G}|$ denotes the number of Gaussians. The weight $w_{\texttt{detail}}^t$ is set to $10^3$ when $t=t^*$ and $1$ otherwise, ensuring consistency between shapes in the canonical shape and at the selected moment $t^*$. Additionally, we enforce a constraint to align the skeleton pose at $t^*$ towards an identity transformation.  {So we define}:
$
L_{\texttt{id}}^{t} = \sum_{j=1}^{|\mathcal{J}|}\|\hat{\mathbf{r}}_j-\mathbf{I}_{q}\|_2^2/|\mathcal{J}|$  
for $t=t^*$ and $L_{\texttt{id}}^{t}=0$ for others, where $\mathbf{I}_{q}$ is the identity quaternion.

%% file: figures_scripts/pipeline.tex
\begin{figure*}[h] 
    \centering
    \includegraphics[width=0.95\textwidth]{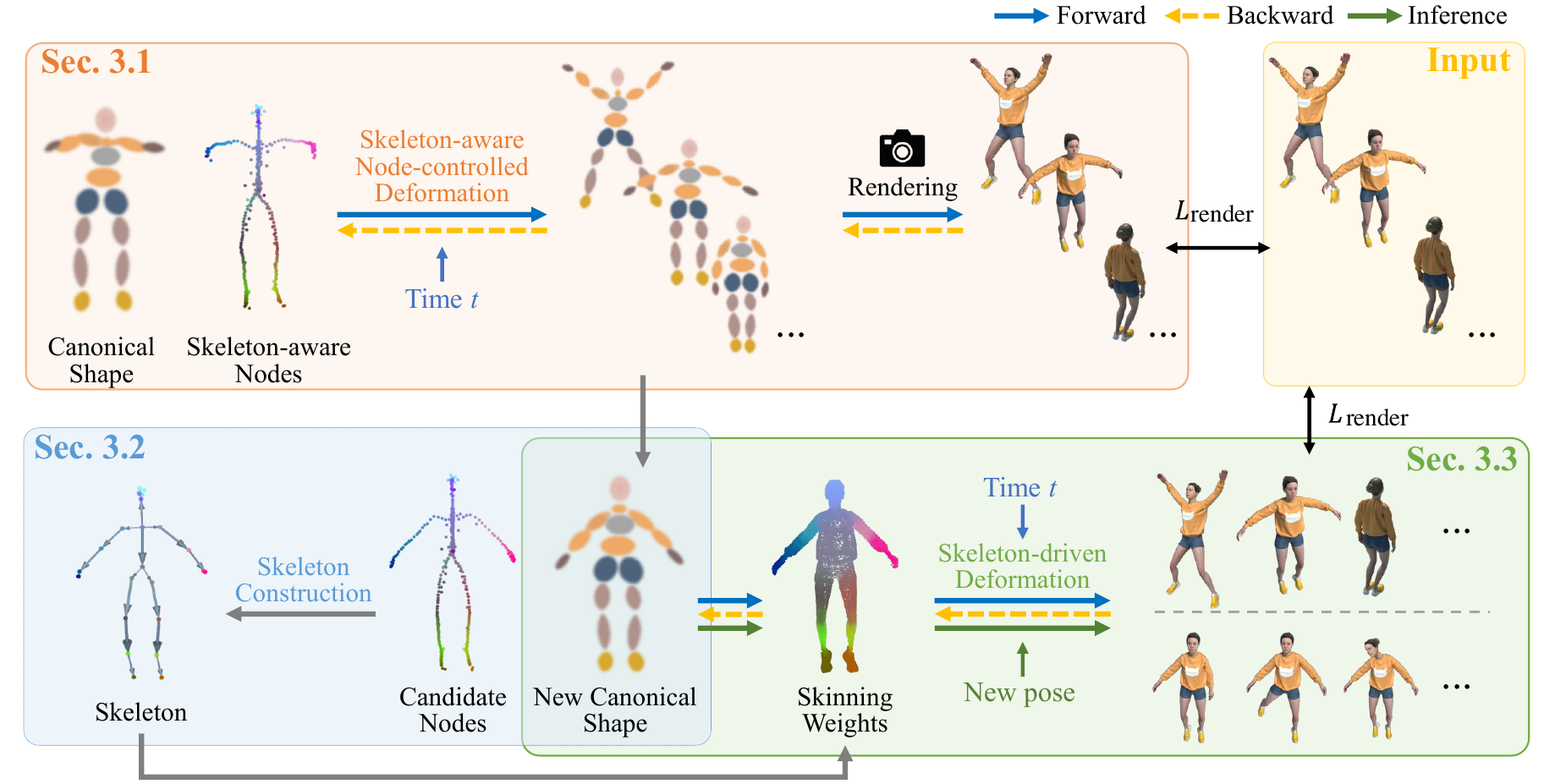} \vspace{-0.2cm}
    \caption{
    Overview of our RigGS. 
    Initially, we construct a canonical 3D Gaussian, coupled with skeleton-aware node-controlled deformation, to begin the 4D reconstruction of the dynamic object. 
    From the resulting skeleton-aware nodes, we then extract a sparse skeleton using a heuristic algorithm. Finally, leveraging the initialized deformation field and 3D Gaussians as starting values, we design learnable skinning weights and optimize a skeleton-driven deformation field. Our RigGS can be utilized for tasks such as editing, interpolation, and motion transfer, enabling real-time high-quality rendering of these new actions.}
    \label{fig:pipeline} 
\end{figure*}

%% file: figures_scripts/skeleton_construction.tex
\begin{figure*}[h] 
    \centering
    \includegraphics[width=0.9\textwidth]{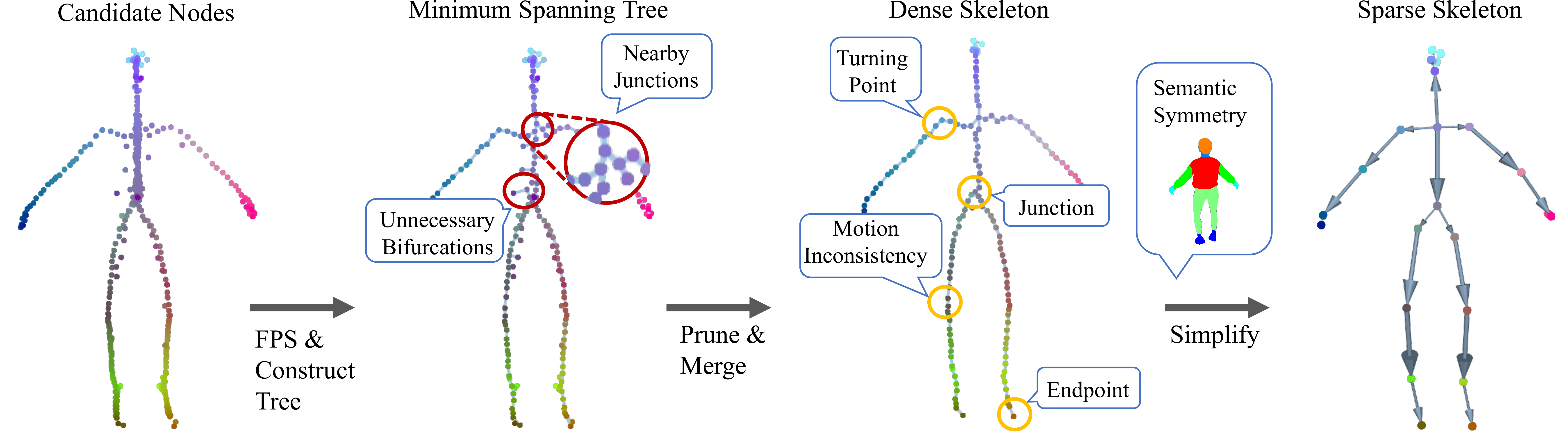}
    \caption{The process of the skeleton construction. The red circles mark the nodes that need to be removed or merged, and the yellow circles mark some locations selected as skeleton joints.
    }
    \label{fig:skeleton_construct} 
\end{figure*}

%% file: sections/results.tex
\section{Results}

\subsection{Experimental Settings}
\mypara{Implementation Details}
In our RigGS, we utilized MLPs with 8 linear layers with a feature dimension of 256 to implement $F_{\Phi}(\cdot)$, $F_{\Psi}(\cdot)$, $F_{\Theta}(\cdot)$, and $F_{\Pi}(\cdot)$.  The ADAM optimizer~\cite{diederik2014adam} was adopted during training. More details are available in the \textit{Supplementary Material}. We conducted all experiments on a single NVIDIA RTX A6000 GPU.

\vspace{0.5em}
\mypara{Datasets}  We used two synthetic datasets: the D-NeRF dataset~\cite{pumarola2021dnerf}, which includes 8 sequences, and the DG-Mesh dataset~\cite{liu2024dynamic}, which includes 6 sequences. Each dataset contains a series of continuous actions. However, the ``Bouncing balls" in the D-NeRF dataset and ``Torus2sphere'' in the DG-Mesh dataset do not match our task setting, and the test view frames in ``Lego" in the D-NeRF dataset do not align with the training actions, as noted in~\cite{huang2024sc}. Therefore, we tested only the remaining 6 sequences from the D-NeRF dataset and 5 sequences from the DG-Mesh dataset. 
Additionally, we considered real-captured datasets: 6 subjects (377, 386, 387,
392, 393, 394) on the ZJU-MoCap dataset~\cite{peng2021neural}.  

\input{figures_scripts/ablas_aniso}
\input{figures_scripts/ablas_projskel}

\vspace{-0.3cm}
\subsection{Ablation Study}
On the D-NeRF dataset, we evaluated the effectiveness of each component of our method using the following variations:
\begin{itemize}
    \item \textit{Anisotropic or isotropic 3D Gaussians.} Although anisotropic 3D Gaussians yield images closer to the ground truth, they suffer significant quality degradation when the new poses differ greatly from the training pose, as shown in Fig.~\ref{fig:ablas_aniso}. Therefore, we use the isotropic 3D Gaussians for all experiments. 
    \item \textit{Without 2D projection loss $L_{\tilde{\texttt{proj}}}^t$ or fixed weight $w_{\tilde{\texttt{proj}}}^t$ during training the skeleton-driven dynamic model.} The 2D projection loss $L_{\tilde{\texttt{proj}}}^t$ assesses how well the skeleton aligns with the 3D Gaussians. We tested performances without this loss and with fixed weights, setting $w_{\tilde{\texttt{proj}}}^t=10^{-3}$.  
    Fig.~\ref{fig:ablas_projskel} shows that without the projection loss, the skeleton cannot be embedded into the 3D Gaussians. Using fixed $w_{\tilde{\texttt{proj}}}^t$ instead of adaptive weights can cause skeletons to protrude beyond the shape in some frames due to inaccuracies in the extracted 2D skeletons. 
    \item \textit{Skeleton construction and skeleton-driven deformation field.} More discussions, numerical results, and visual results are available in the \textit{Supplementary Material}.
\end{itemize}

\subsection{Comparisons}
\mypara{Novel View Synthesis}
To demonstrate the advantages of our RigGS, we compared it with state-of-the-art approaches: D-NeRF~\cite{pumarola2021dnerf}, TiNeuVox~\cite{fang2022TiNeuVox}, 4D-GS~\cite{wu20244dgs}, SC-GS~\cite{huang2024sc}, and AP-NeRF~\cite{uzolas2024template} in novel viewpoint synthesis\footnote{Since D-NeRF failed to run on the "beagle" sequence on DG-Mesh dataset, the average results of the other four datasets are presented.}. Except for AP-NeRF, other methods do not bind the skeleton, and \textit{do not} support editing the object through repose. 
As shown in Table~\ref{Tab:compare_nvs}, our rendering accuracy is slightly lower than SC-GS but significantly higher than the other methods. Besides the use of anisotropic 3D Gaussians, SC-GS achieves better detail fitting due to its more freedom of deformation field (512 control points) and temporally adjustments in scales and rotations of 3D Gaussians. 
However, these configurations do not apply to our task because we require sparse skeletons and require all deformation variables to be pose-related for generating plausible movements.  Fig.~\ref{fig:compare_nvs} shows that visually, our method is comparable to SC-GS without significant disadvantages.
Additionally, we also compared our method with AP-NeRF on the real-captured ZJU-MoCap dataset, and show the numerical results in Table~\ref{Tab:compare_zju}.
More visual results are available in the \textit{Supplementary Material}.

\input{table_scripts/compare_blender_dgmesh}

\input{figures_scripts/compute_nvs}

\input{figures_scripts/compare_skeleton}

\input{figures_scripts/pose_transfer}

\input{table_scripts/compare_zju}

\vspace{0.5em}
\mypara{Skeleton and Skinning Weights} 
Closely related to our research, AP-NeRF~\cite{uzolas2024template} stands as an automated framework designed for the construction of skeletons and the modeling of articulated objects. Initialized with TiNeuVox~\cite{fang2022TiNeuVox}, it utilizes reconstructed surface points to extract a 3D skeleton employing the Medial Axis Transform (MAT), refining the skeleton progressively through optimization steps. In contrast, our approach diminishes the dependency on the quality of 3D reconstruction.
As shown in Fig.~\ref{fig:compare_skeleton}, we compared AP-NeRF with our method on the constructed skeleton and skinning weights. Furthermore, we manipulated the poses of the skeletons to generate the same new actions for the object. We can observe that our skeleton is more reasonable, and the skinning weights are smoother, avoiding discontinuities at the joints. In addition, we have higher rendering quality.

\vspace{0.5em}
\mypara{Editing and Interpolation}
Thanks to our skeleton-driven deformation field, we can create new motions for the object by editing the rotations of the skeleton bones. To facilitate this process, we have developed a GUI that allows for interactive editing and real-time rendering of the results. Figs.~\ref{fig:teaser},~\ref{fig:compare_skeleton} and~\ref{fig:edit_interpolate} display several editing performances.  Additionally,  as shown in Fig.~\ref{fig:edit_interpolate}, we can easily interpolate between two poses of an object in a plausible manner.

\input{figures_scripts/edit_interpolate}

\vspace{0.5em}
\mypara{Motion Transfer}
Our method can also be utilized for motion transfer tasks. As illustrated in Fig.~\ref{fig:pose-transfer}, for two sequences with similar structures, we first manually annotate the correspondences between their skeletons, repose the target object to match the source object's pose, and then transfer the source object's pose sequences to drive the target object, generating new motion sequences.

%% file: figures_scripts/ablas_aniso.tex
\begin{figure}[h] 
    \centering
\includegraphics[width=1\columnwidth]{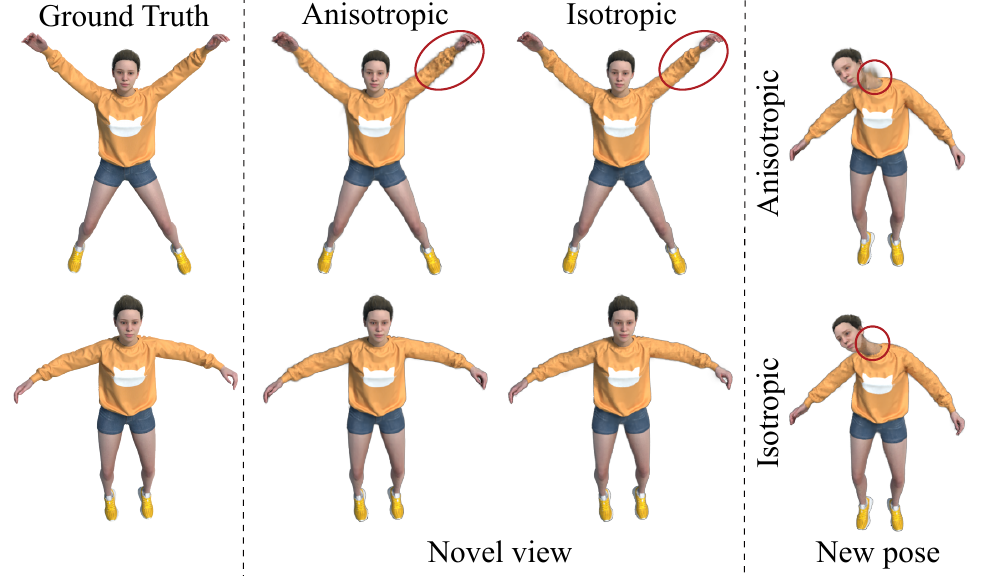}
    \vspace{-0.3cm}
    \caption{Novel view rendering and repose results via anisotropic or isotropic 3D Gaussians. }
    \label{fig:ablas_aniso} 
\end{figure}

%% file: figures_scripts/ablas_projskel.tex
\begin{figure}[h] 
    \centering
    \includegraphics[width=0.95\columnwidth]{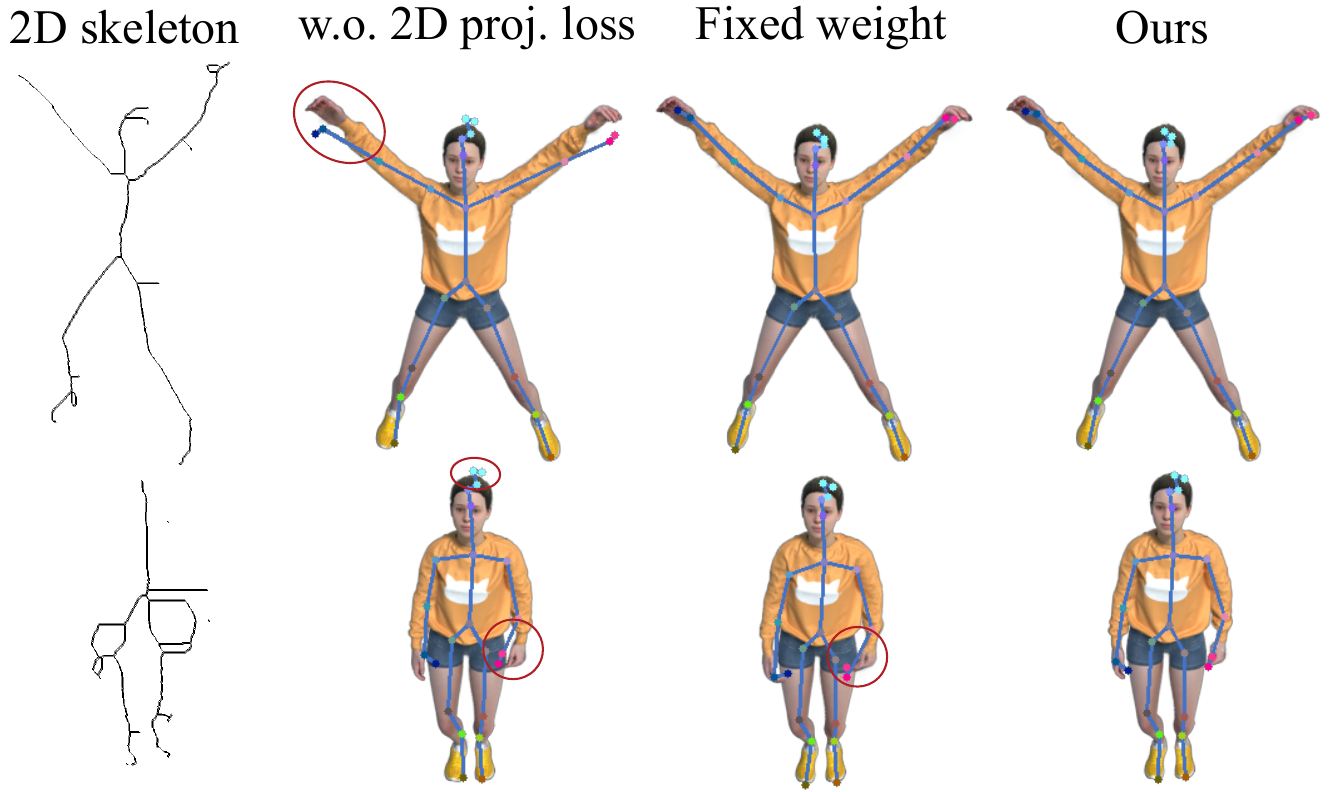}
    \vspace{-0.3cm}
    \caption{Comparison of visual results for different variants on 2D projection loss term.}
    \label{fig:ablas_projskel} 
\end{figure}

%% file: table_scripts/compare_blender_dgmesh.tex
\begin{table}[t]
	\caption{Comparisons of the average precision (PSNR $\uparrow$  /  SSIM $\uparrow$ / LPIPS $\downarrow$) on the D-NeRF dataset and DG-Mesh dataset.}\vspace{-0.2cm}
	\label{Tab:compare_nvs}
		\setlength{\tabcolsep}{5.3pt}
	\centering
	\begin{footnotesize}
		\begin{tabular}{ c   c  c }
            \toprule 
            \multirow{2}{*}{Method}  & \multicolumn{1}{c}{D-NeRF~\cite{pumarola2021dnerf}} & \multicolumn{1}{c}{DG-Mesh~\cite{liu2024dynamic}}
            \\ 
             &  PSNR /  SSIM / LPIPS & PSNR /  SSIM / LPIPS \\
            \midrule
            
            D-NeRF~\cite{pumarola2021dnerf} & 30.48 / 0.973 / 0.0492 & 28.17 / 0.957 / 0.0778 \\
            TiNeuVox~\cite{fang2022TiNeuVox} &  32.60 / 0.983 /	0.0436 & 31.95 / 0.967 / 0.0477 \\
            4D-GS~\cite{wu20244dgs} &  33.25 /	0.989 / 0.0233 & 33.96 / 0.979 / 0.0272 \\
            SC-GS~\cite{huang2024sc} &  \textbf{43.04} / \textbf{0.998} / \textbf{0.0066} & \textbf{38.96} / \textbf{0.993} / \textbf{0.0136} \\
            \midrule 
            AP-NeRF~\cite{uzolas2024template} &  30.94 / 0.970 / 0.0350 & 31.83 / 0.967 / 0.0460 \\
            Ours &  \underline{40.82} / \underline{0.996} / \underline{0.0112} &\underline{37.65} / \underline{0.991} /	\underline{0.0169} \\
            \bottomrule
		\end{tabular}
	\end{footnotesize}
\end{table}

%% file: figures_scripts/compute_nvs.tex
\begin{figure}[t] 
    \centering
\includegraphics[width=\columnwidth]{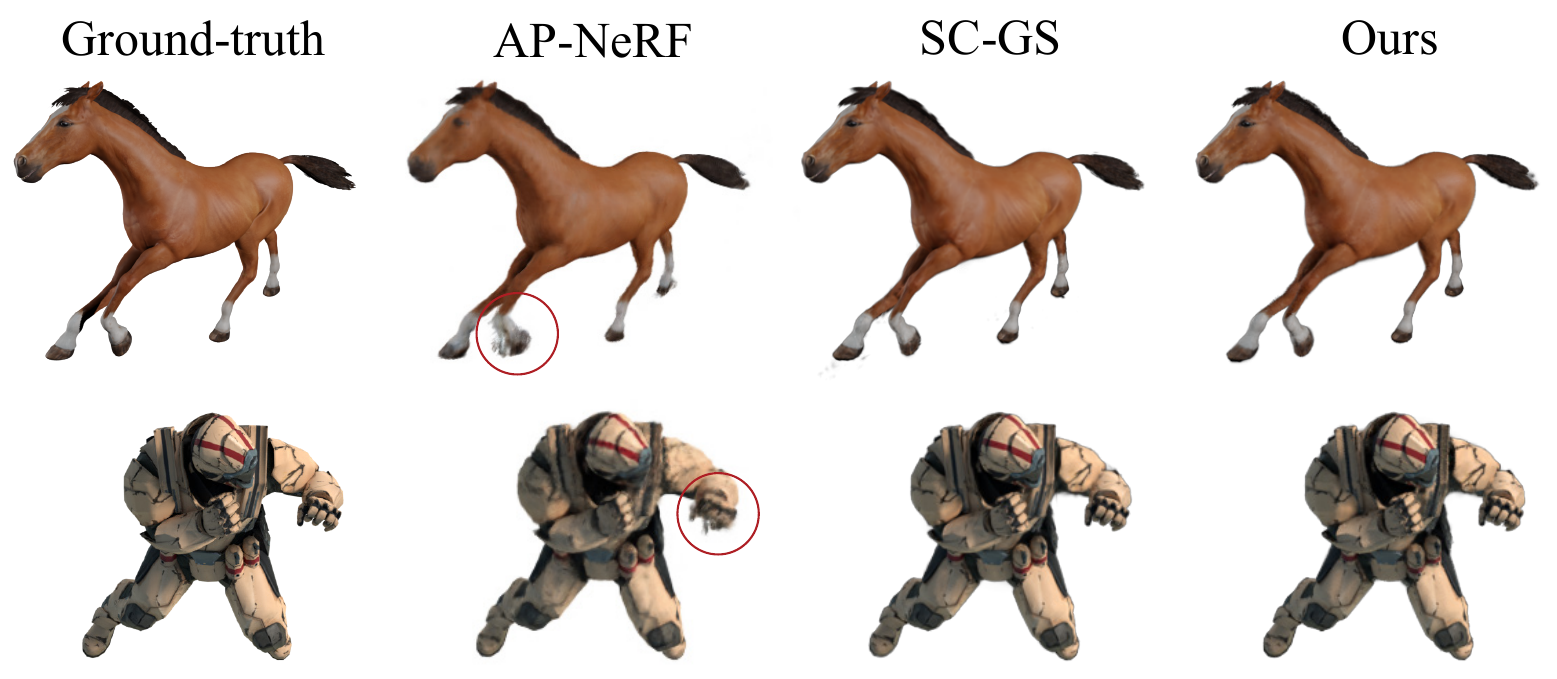}
    \vspace{-0.6cm}
    \caption{Comparison with state-of-the-art methods for novel view synthesis.}
    \label{fig:compare_nvs} 
\end{figure}

%% file: figures_scripts/compare_skeleton.tex
\begin{figure}[b] 
    \centering
    \includegraphics[width=\columnwidth]{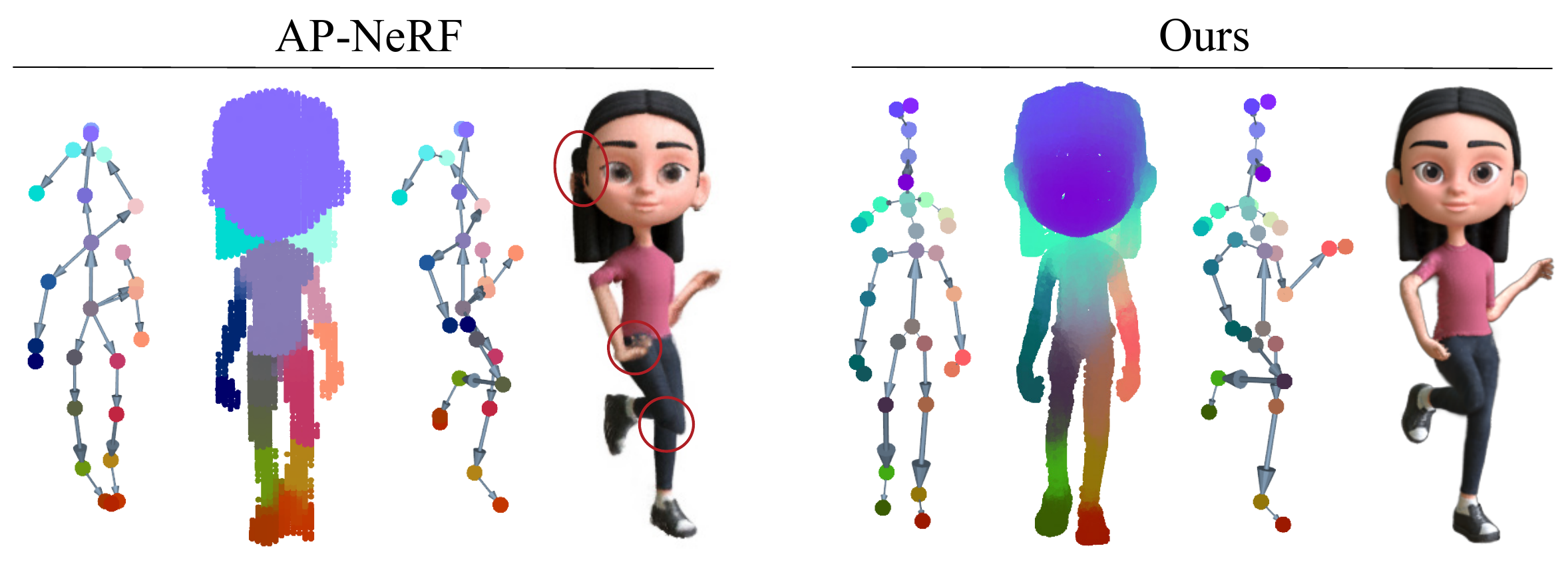}
    \caption{Comparisons of the skeleton, the skinning weights, edited skeleton pose, and the corresponding reposed objects.}
    \label{fig:compare_skeleton} 
\end{figure}

%% file: figures_scripts/pose_transfer.tex
\begin{figure*}[h] 
    \centering
    \includegraphics[width=0.95\textwidth]{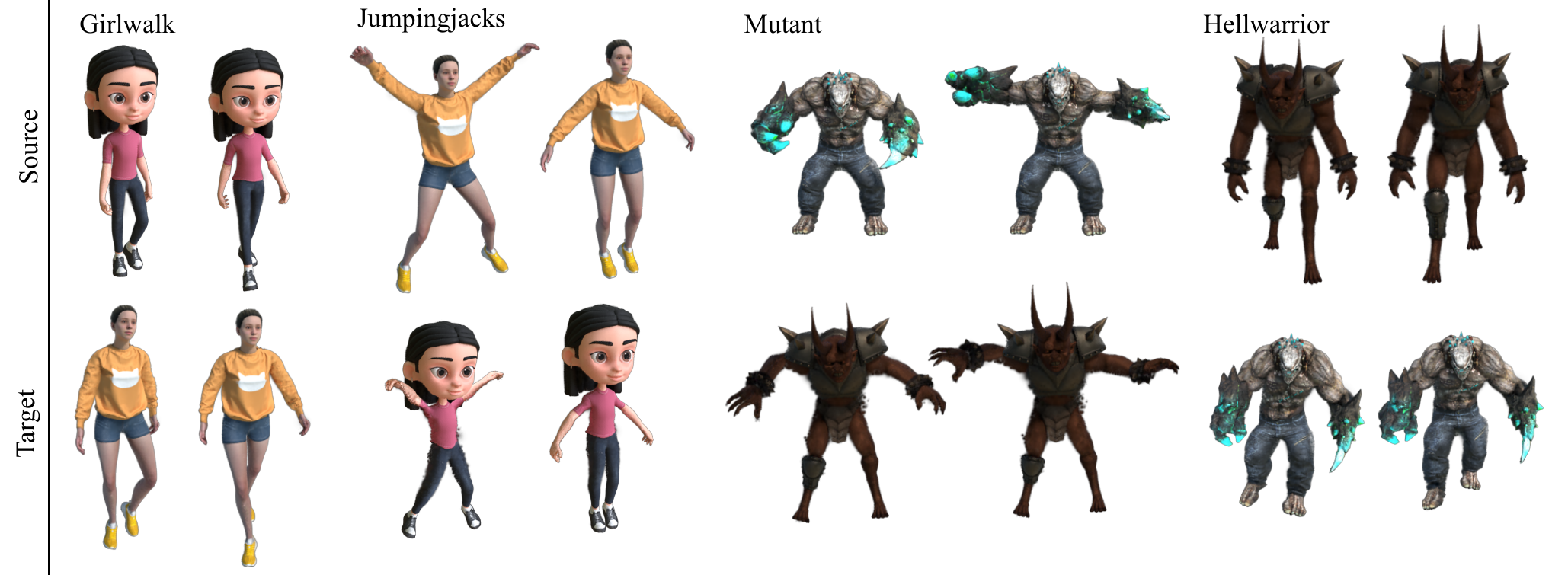}
    \caption{Visualization of motion transfer from the source object to the target object.}
    \label{fig:pose-transfer} 
\end{figure*}

%% file: table_scripts/compare_zju.tex
\begin{table}[t]
	\caption{Comparisons of the average precision on ZJU-MoCap. 
    }\vspace{-0.25cm}
	\label{Tab:compare_zju}
		\setlength{\tabcolsep}{5.3pt}
	\centering
	\begin{footnotesize}
		\begin{tabular}{ c   c  c  c}
            \toprule 
            Method  & PSNR $\uparrow$ & SSIM $\uparrow$ & LPIPS $\downarrow$ \\ 
            \midrule
            AP-NeRF~\cite{uzolas2024template} & 25.62 & 0.919 & 0.0934 \\
            Ours & 33.54 & 0.975 & 0.0327 \\
            \bottomrule
		\end{tabular}
	\end{footnotesize}
    \vspace{-0.35cm}
\end{table}

%% file: figures_scripts/edit_interpolate.tex
\begin{figure}[h] 
    \centering
    \includegraphics[width=\columnwidth]{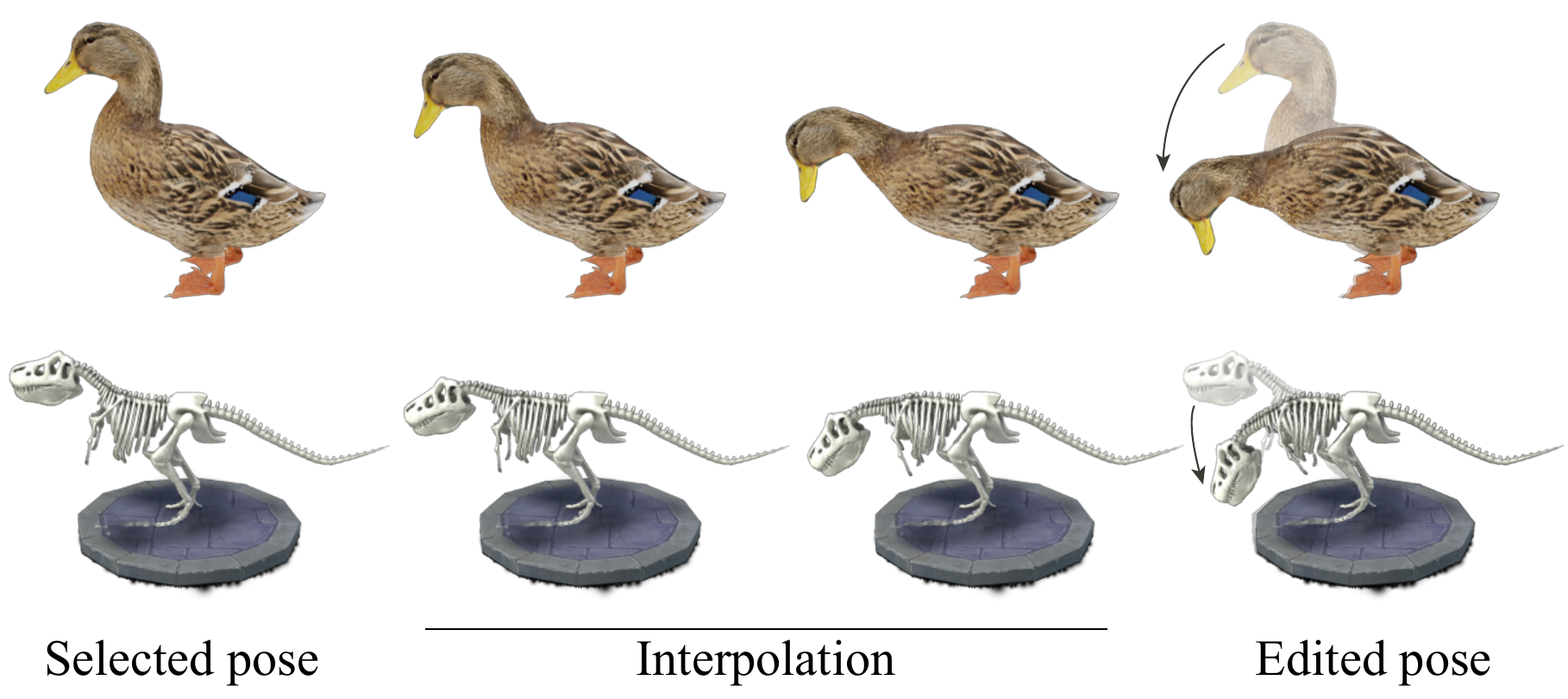}
    \caption{Visualizations of the selected pose, the edited pose, and the interpolated results between them.}
    \label{fig:edit_interpolate} 
\end{figure}

%% file: sections/conclusion.tex
\section{Conclusion and Discussion}

We have presented RigGS, a skeleton-driven modeling approach that reconstructs articulated objects from 2D videos without relying on any template priors. First, we initialized the reconstruction of dynamic objects from input videos using a skeleton-aware node-controlled deformation field combined with a canonical 3D Gaussian representation, which also yields candidate points for the skeleton. Second, we introduced an automated skeleton extraction algorithm to obtain a sparse skeleton from these candidate points. Finally, we established a skeleton-driven dynamic modeling approach that binds the canonical 3D Gaussians with the skeleton, enabling tasks such as editing, interpolation, and motion transfer, while rendering high-quality images from novel viewpoints. Experimental results demonstrated that our method achieves rendering quality close to state-of-the-art novel view synthesis methods while easily generating new poses for the objects.

While our RigGS has shown promising results in many cases, its effectiveness may be limited when dealing with sparse viewpoints, inaccurate estimation of camera poses, or excessive motion. The pose-related appearance is also not modeled. More useful techniques for 3D reconstruction and editing will be helpful in solving these challenging problems. Additionally, exploring more complex input signals, such as text and images, for semantic skeleton pose editing could be an interesting future direction.

%% file: sections/appendix.tex
\maketitlesupplementary

\section{Details of Skeleton Construction}
We provide a detailed description of the skeleton construction process. Alg.~\ref{Alg:skeleton-extraction} demonstrates the overall process.

\vspace{1em}
\mypara{Prune or Merge for Dense Skeleton Construction} 
After constructing the minimum spanning tree, due to the presence of noise, we need to remove redundant branches and merge closely located joints. Specifically, if an endpoint passes through fewer than $r$ connected points on its way to the nearest junction, we consider it as an unnecessary point and remove this endpoint along with these connection points.
In addition, we will merge two junctions if the number of connection points between them is less than $r$, and remove these connected points if they exist.
Since the distribution of these nodes is uniform, such operations generally do not remove important feature points with long neighboring edges. By default, we set $r=3$. 

\vspace{1em}
\mypara{Construction of $\mathcal{J}$ for Skeleton Simplification}
The following steps are performed to obtain the joint set $\mathcal{J}$ of the sparse skeleton:

\vspace{0.5em}
\noindent\textbf{\textit{Step 1: Initialize $\mathcal{J}$ and define initial paths}.} We regard all junctions and endpoints as key points of $\mathcal{J}_d$ and add them into $\mathcal{J}$. For any two connected key points $(\mathbf{J}_a,\mathbf{J}_b)$, i.e. when moving from $\mathbf{J}_a$ to $\mathbf{J}_b$, there are no other key points, we define a path $\texttt{P}_{ab}=\{\mathbf{J}_a, \mathbf{p}_1^{ab},...,\mathbf{p}_m^{ab},\mathbf{J}_b\}$ (Fig.~\ref{fig:select_key_joints} (a)) where $\mathbf{p}_i^{ab}\in\mathcal{J}_d$ are the passing points from $\mathbf{J}_a$ to $\mathbf{J}_b$.  

\input{figures_scripts/select_key_joints}

\input{algorithm_scripts/construct_skeleton}

\vspace{0.5em}
\noindent\textbf{\textit{Step 2: Select a candidate joint from each path}.} From these passing points $\{\mathbf{p}_i^{ab}\}$, we select a point that is more likely based on geometric position and motion information as the candidate points of $\mathcal{J}$. 
We assume that geometric turning points are more likely to be joint points. 
So we first connect $\mathbf{J}_a$ and $\mathbf{J}_b$ with a straight line segment $l_{ab}$ and calculate the distance $D(\mathbf{p}_{i}^{ab}, l_{ab})$ from $\mathbf{p}_i^{ab}$ to $l_{ab}$. See Fig.~\ref{fig:select_key_joints} (b), the point with the maximum distance is more likely to be a geometric turning point.
However, due to noise in $\mathcal{J}_d$, directly selecting points based on this distance may lead to selecting points very close to $\mathbf{J}_a$ or $\mathbf{J}_b$ (such as red points in Fig.~\ref{fig:select_key_joints} (c)), which are undesirable joints. Therefore, we define the following score
\begin{equation}
\label{eq:distance_score}
\texttt{s}_i^{ab} = D(\mathbf{p}_i^{ab},l_{ab}) - 0.1\min(D(\mathbf{p}_i^{ab},\mathbf{J}_a), D(\mathbf{p}_i^{ab},\mathbf{J}_b)),
\end{equation}
where $D(\mathbf{p}_i^{ab}, \mathbf{J}_a)$ and $D(\mathbf{p}_i^{ab}, \mathbf{J}_b)$ are the distance between $\mathbf{p}_i^{ab}$ with $\mathbf{J}_a$ and $\mathbf{J}_b$, separately. 

To consider motion information, at each time $t$, we transform $\mathbf{p}_i^{ab}, \mathbf{J}_a$, and $\mathbf{J}_b$ to the position of the ${\mathbf{p}_i^{ab}}^t, \mathbf{J}_a^t$, and $\mathbf{J}_b^t$ according to the initial deformation field in Sec.~3.1 and compute ${\texttt{s}_i^{ab}}^t$ according to Eq.~\eqref{eq:distance_score}. The point $\mathbf{p}_*^{ab}$ with the maximal $\sum_{t}{\texttt{s}_i^{ab}}^t$ is selected as the candidate joint. 

\vspace{0.5em}
\noindent\textbf{\textit{Step 3: Add paths and continue to select candidate points}.} 
The new candidate point $\mathbf{p}_*^{ab}$ and endpoints $\mathbf{J}_a$ and $\mathbf{J}_b$ form two new paths. Repeat the \textit{step 2} to find new candidate points. 
Fig.~\ref{fig:select_key_joints} shows this process.

\input{algorithm_scripts/selection_candidate}

\vspace{0.5em}
\noindent\textbf{\textit{Step 4: Enhance the symmetry of candidate joints using precomputed DINOv2 features}~\cite{oquab2023dinov2}.}
Considering that many objects exhibit symmetry, we further utilize semantic information to enhance the symmetry of candidate joints selected by \textit{step 2-3}. For two paths $\texttt{P}_{ab}$ and $\texttt{P}_{ef}$ in \textit{step 1}, we consider they are symmetric if their lengths and semantic labels are similar. 
By projecting $\mathbf{J}_a^t, \mathbf{J}_b^t$ and ${\mathbf{p}_i^{ab}}^t$ onto the image according to the camera perspective, we can obtain semantic classification based on image segmentation using DINO features.
We set the semantic label of $\mathbf{J}_a, \mathbf{J}_b$ and ${\mathbf{p}_i^{ab}}$ as the median of these labels at all times to represent the semantic category of the majority of frames.
Since these semantic labels are also not precise, we count the number of different kinds $\texttt{sem}(\texttt{P}_{ab})$ and $\texttt{sem}(\texttt{P}_{ef})$ of semantics that appear on $\texttt{P}_{ab}$ and $\texttt{P}_{ef}$ respectively. If $|\texttt{sem}(\texttt{P}_{ab})\cap \texttt{sem}(\texttt{P}_{ef})| > \epsilon_{2}\cdot |\texttt{sem}(\texttt{P}_{ab})\cup \texttt{sem}(\texttt{P}_{ef})|$, then we consider them to be similar in semantics.

If $\texttt{P}_{ab}$ and ${\texttt{P}_{ef}}$ are similar in length and semantics, then we will perform symmetrical correction on them so that the number and distribution of joints along the two paths are similar. 
We tend to choose the path with a moderate number of candidate points as templates (closer to $\epsilon_3$) and adjust the other path to be similar. 
Without loss of generality, let's assume $\texttt{P}_{ab}$ is the template. we will discard the candidate joints in $\texttt{P}_{ef}$ and reselect points from $\texttt{P}_{ef}=\{\mathbf{J}_e, \mathbf{p}_{d_1}, ...,\mathbf{p}_{d_{n_{ab}}}, \mathbf{J}_f\}$, where $\mathbf{p}_{d_i}$ is selected by $\frac{\texttt{len}([\mathbf{J}_e, \mathbf{p}_{d_i}])}{\texttt{len}([\mathbf{J}_e, \mathbf{J}_f])}\approx \frac{\texttt{len}([\mathbf{J}_a, \mathbf{p}_{i}^{ab}])}{\texttt{len}([\mathbf{J}_a, \mathbf{J}_b])}$. 
Here $\texttt{len}(x,y)$ denotes the length of the path between $x$ and $y$.  
Based on our experience, setting $\epsilon_{1}=30\%,\epsilon_2=60\%,\epsilon_3=3$ is a robust choice. 
After processing each path pair, we obtain the final sparse joint set $\mathcal{J}$.

\input{figures_scripts/ablas_off_skinmlp}

\section{Implementation Details}
During the training process, we first trained the initialization stage for 80,000 iterations, optimizing 3D Gaussians $\mathcal{G}$, the positions of skeleton-aware nodes and the parameters of MLP $F_{\Theta}$ for the corresponding time-related rotations and translations. Then we obtained the initialized deformation field and candidate nodes of skeleton significance. We set the weight $w_{\texttt{proj}}=10^{-3}$, and dynamically decreased $w_{\texttt{arap}}$ from $10^{-4}$ to $0$ during iterations.
After completing the initialization training, we obtained a new canonical shape and constructed the skeleton. The new canonical shape will serve as the initial 3D Gaussian representation for the skeleton-driven dynamic model. We obtained the initial values of skeleton-driven deformation by pointwise supervision with the skeleton-aware node-controlled deformation field.  Finally, we performed the training process for 100,000 iterations, optimizing 3D Gaussians $\mathcal{G}$, the parameters of MLP $F_{\Phi}$ for time-related skeleton poses and translations, scaling factors $\{\eta_{i,j}\}_{j=2}^{\mathcal{J}}$ and the parameters of $F_{\Psi}$ for learnable skinning weights, as well as the parameters of $F_{\Pi}$ for pose-dependent detail deformation. At this stage, the positions of joints and parent indices of the skeleton are fixed, as treating them as variables can lead to instability. During the inference stage, we solely executed the skeleton-driven deformation model.

\section{More Ablation Studies}
Table~\ref{Tab:ablation} lists the numerical results of our ablation experiments. Additionally, we validated the effectiveness of our skeleton-driven deformation module. We compared our method without MLP $F_{\Phi}(\gamma_t(t))$ for skinning weight or without pose-dependent detail deformation. From Table~\ref{Tab:ablation} and Fig.~\ref{fig:ablas_off_skinmlp}, we can see these variants result in a slight decrease in rendering quality, indicating that these two modules are effective in matching details without changing the main parts of the deformation. 
\input{table_scripts/ablas}

\input{figures_scripts/ablation_motion}

\vspace{0.1cm}
\mypara{Robustness of Skeleton Construction under Large Motion}
To evaluate the robustness of our method under large motion, we selected the sequence ``393" from the ZJU-MoCap dataset. We tested the sequence using frames indexed as $[0,k,2k,...,nk]$, where a larger value of $k$ corresponds to greater motion.  We set $k=1,5,10$ and $n=65$ and show the results in Fig.~\ref{fig:ablas_large_motion}, which demonstrates the robustness of our skeleton extraction method in handling large motions.

\section{More results}

\input{figures_scripts/supp_scgs_edit}

\input{figures_scripts/supp_compare_zju}
\input{table_scripts/supp_compare_zju}

\input{table_scripts/supp_compare_blender}

\mypara{More Results for Editing}
SC-GS~\cite{huang2024sc} introduces sparse control points and ARAP regularization, allowing them to complete editing tasks by fixing some control points and moving other points. As shown in Fig.~\ref{fig:supp_scgs_edit}, using SC-GS to edit the object is not easily controllable. For example, we want only to edit the arm, but the leg is also moving. Furthermore, even by adding more fixed points, achieving reasonable edits of movements is still challenging. As demonstrated in Figs.~\ref{fig:supp_skeleton} and \ref{fig:supp_skeleton_dgmesh}, we also compared the extracted skeletons, skinning weights, and editing performance of our RigGS with those of AP-NeRF~\cite{uzolas2024template} on more examples.  Due to the different canonical shapes established by the AP-NeRF and ours, the poses of the skeletons / skinning weights are different.
We aligned the skeletons generated by two methods to the same pose to create new poses for the objects. It can be observed that our method generally yields more reasonable results with higher clarity. 
More edited animations are shown in the \textbf{Video Demo}.

\mypara{More Details on the ZJU-MoCap Dataset}
We conduct more experiments on the real-captured dataset, ZJU-MoCap dataset~\cite{peng2021neural}, and show the comparisons with AP-NeRF. 
Since our template-free method performs reconstruction and rigging simultaneously, it faces challenges with videos captured by a fixed camera. Improved results can be achieved when the camera is allowed to move. Therefore, we used 6 cameras (1, 5, 9, 13, 17, 21) to simulate monocular videos with camera movement. At each time, we only select one image captured by one of these cameras. Additionally, to more accurately capture the motion of the human body, we use frame-by-frame corresponding SMPL vertices to initialize the 3D Gaussians and deformation fields.
We show the comparisons on the other 17 cameras in Table~\ref{Tab:supp_compare_zju} and Fig.~\ref{fig:compare_zju}. We can see that the performance of our method is significantly better than AP-NeRF. 

\vspace{0.3em}
\mypara{More Results for Novel View Synthesis} 
We present the numerical results for each sequence in Table~\ref{Tab:comparisons-dnerf} and Table~\ref{Tab:comparisons-dgmesh}. Except for SC-GS~\cite{huang2024sc}, we can see our rendering quality is significantly better than that of other methods. Additionally, we showcase more visual results in Fig.~\ref{fig:supp_nvs} and the complete motion sequences in the \textbf{Video Demo.} Despite SC-GS having higher numerical accuracy, from Fig.~\ref{fig:supp_nvs}, we can see that it exhibits more artifacts compared to our method, such as in the wings of the ``Bird'' and the legs of the ``Horse''.
Furthermore, when dealing with real data, the performance of SC-GS is notably inferior to our method, as shown in the \textbf{Video Demo}. 

\section{Failure Cases of Skeleton Construction}
Our method is dependent on the quality of 2D skeleton extraction and silhouettes. Although we proposes an adaptive weighting mechanism to mitigate the impact of erroneous 2D skeletons, when a significant portion of frames contain inaccurate estimations, our method fails to produce semantically plausible skeletal structures (Fig.~\ref{fig:supp_failure_case} (a)). Developing higher-quality 2D skeleton extraction methods will greatly improve the quality of the resulting skeleton tree. Secondly, since our automated rigging is inherently related to motions, when two adjacent regions exhibit no relative transformations over the input sequence, our method struggles to distinguish them, resulting in their inability to be properly separated (Fig.~\ref{fig:supp_failure_case} (b)). Integrating semantic segmentation or similar techniques to model the skeleton and skinning weights represents a promising direction for future research.
\input{figures_scripts/supp_failure_case}

\input{figures_scripts/supp_skeleton}
\input{figures_scripts/supp_skeleton_dgmesh}
\input{figures_scripts/supp_nvs}
\input{table_scripts/supp_compare_dgmesh}

%% file: figures_scripts/select_key_joints.tex
\begin{figure}[h] 
    \centering
    \includegraphics[width=\columnwidth]{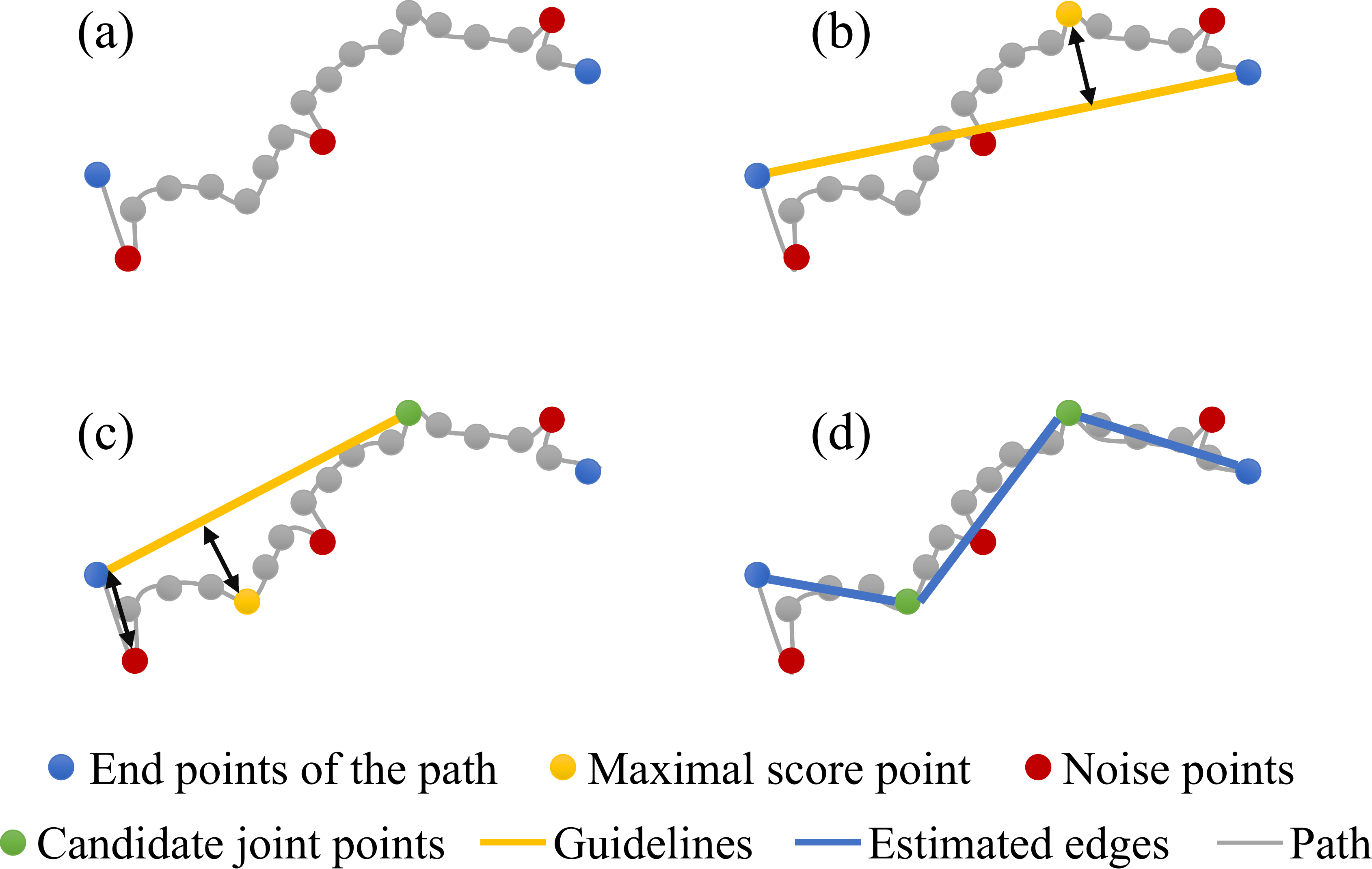}
    \caption{Selection of candidate joints.}
    \label{fig:select_key_joints} 
\end{figure}

%% file: algorithm_scripts/construct_skeleton.tex
\begin{algorithm}[t]
	\caption{Construction of skeleton.}
	\label{Alg:skeleton-extraction}
	\KwIn{Skeleton candidate nodes $\{\mathbf{c}^{t^*}\}$ and the corresponding semantic labels\;
   }
	\KwResult{Skeleton including joints $\mathcal{J} = \{\mathbf{J}_j\}$ and parent indices $\{\mathbf{J}_{A(j)}\}$.} 
	\BlankLine
    \tcp{Construct dense skeleton.}
    Perform farthest point sampling to obtain uniformly distributed control points $\{\mathbf{c}_s^{t^*}\}_{s\in \mathcal{S}}$\;
    Construct the edge set $\mathcal{E}$ for $\{\mathbf{c}_s^{t^*}\}$ using Prim's algorithm\;
    Prune redundant endpoints and merge close junctions\;
    \tcp{Skeleton simplification.}
    $\mathcal{J} = \emptyset$\;
    Add all junctions and ends to $\mathcal{J}$\;
    Establish paths $\{\texttt{P}\}$ between two connected nodes in $\mathcal{J}$\;
    \For{$\forall (J_a,J_b)\in\mathcal{J}$}
    {
      Compute the candidate joints set $h_{ab}$ according to Alg.~\ref{Alg:candidate-joints}\;
    }
    \tcp{Symmetry enhancement based on semantic labels.}
    \For {$\forall (\texttt{P}_{ab},\texttt{P}_{ef})$}
    {
     \If{$\left||\texttt{P}_{ab}|-|\texttt{P}_{ef}|\right|<\epsilon_1\cdot\max(|\texttt{P}_{ab}|, |\texttt{P}_{ef}|)$~~and~~$|\texttt{sem}(\texttt{P}_{ab})\cap \texttt{sem}(\texttt{P}_{ef})| > \epsilon_{2}\cdot |\texttt{sem}(\texttt{P}_{ab})\cup \texttt{sem}(\texttt{P}_{ef})|$}
     {
        Select a path with more moderate number of candidate joints\;
        Reselect the candidate joints for other path according to the selected path\;
     }
     Add these candidate joints into $\mathcal{J}$\;
    }
\end{algorithm}

%% file: algorithm_scripts/selection_candidate.tex
\begin{algorithm}[t]
{
	\caption{Selection of candidate joints on a path.}
	\label{Alg:candidate-joints}
	\KwIn{$\texttt{P}_{ab}=[\mathbf{J}_a,\mathbf{p}_1,...,\mathbf{p}_m,\mathbf{J}_b]$, current candidate joints set $h_{ab}$\;
   }
	\KwResult{Updated candidate joints set $h_{ab}$.} 
	\BlankLine
     Set the candidate joints set $h_{ab}=\emptyset$ for $\texttt{P}_{ab}$\;

     Compute ${\texttt{s}_i^{ab}}^t$ according to Eq.~\eqref{eq:distance_score}\;
    $\mathbf{p}_*^{ab} = \arg\max_{\mathbf{p}_i}\sum_{t}{\texttt{s}_i^{ab}}^t/|\mathcal{I}|$\;
     \eIf{$\sum_{t}D_t(\mathbf{p}_*^{ab}, l_{ab})/|\mathcal{I}| < \overline{l}$}
     {
            \Return $h_{ab}$\;
    }
    {
        $h_{ab} = h_{ab} + \{\mathbf{p}_*^{ab}\}$\;
        Perform Alg.~\ref{Alg:candidate-joints} with $[\mathbf{J}_a,\mathbf{p}_1,...,\mathbf{p}_*^{ab}]$ and $h_{ab}$ as input\;
        Perform Alg.~\ref{Alg:candidate-joints} with $[\mathbf{p}_*^{ab},...,\mathbf{J}_b]$ and $h_{ab}$ as input\;
        \Return $h_{ab}$\;
    }
}
\end{algorithm}

%% file: figures_scripts/ablas_off_skinmlp.tex
\begin{figure*}[h] 
    \centering
    \includegraphics[width=1\textwidth]{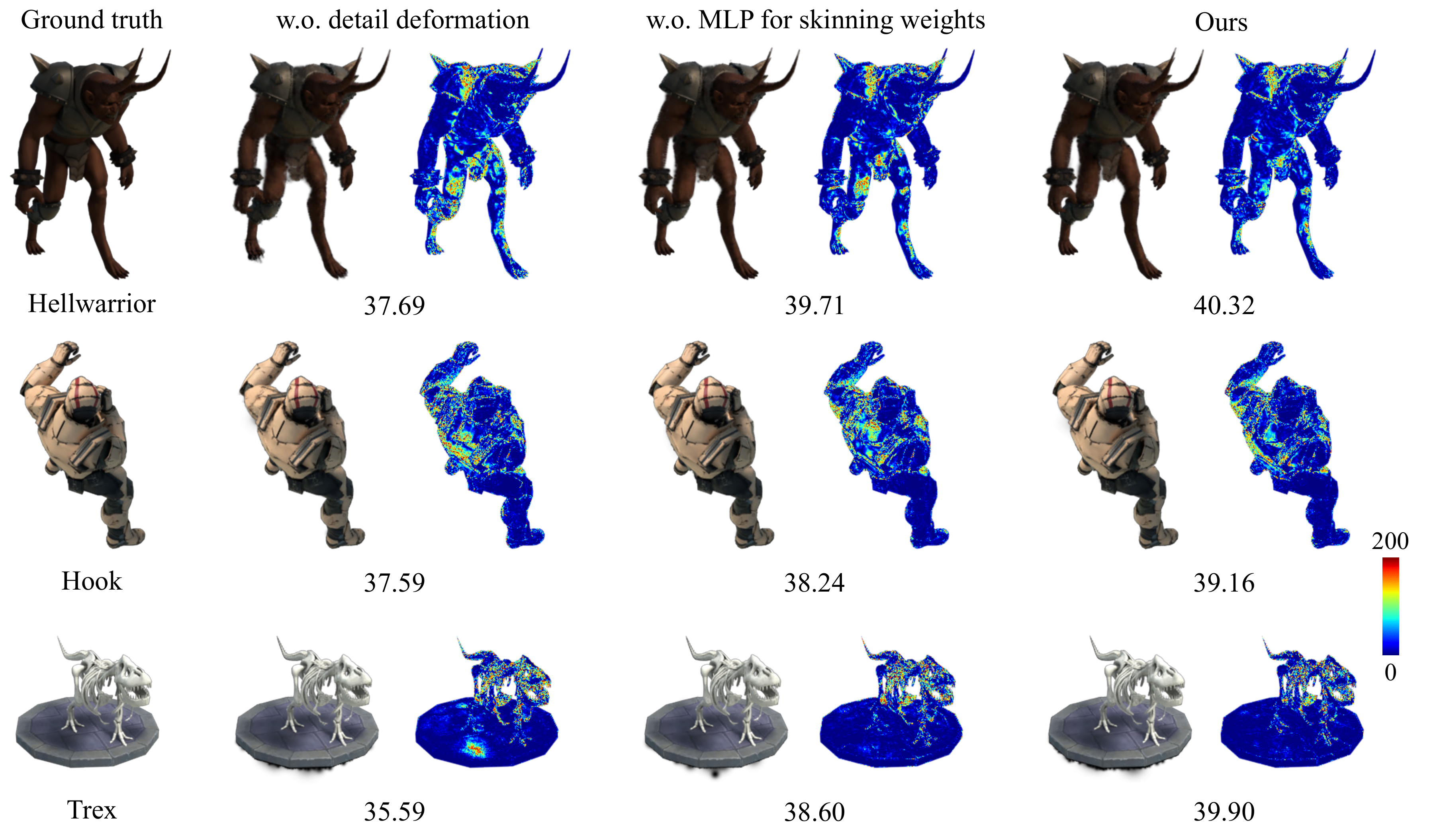}
    \caption{Visual comparisons of our method, without MLP for skinning weights or pose-dependent detail deformation, with annotated PSNR values.}
    \label{fig:ablas_off_skinmlp} 
\end{figure*}

%% file: table_scripts/ablas.tex
\begin{table}[h]
	\caption{Ablation studies of our method on the D-NeRF dataset~\cite{pumarola2021dnerf}. }
	\label{Tab:ablation}
		\setlength{\tabcolsep}{3pt}
	\centering
	\begin{small}
		\begin{tabular}{ c | c c c}
            \toprule 
            Variants & PSNR $\uparrow$  & SSIM $\uparrow$ & LPIPS $\downarrow$ \\
            \midrule
            
            w.o. 2D proj. loss $L_{\tilde{\texttt{proj}}}^t$ & 40.60 &	0.996 &	0.0108 \\
            Fixed weight $w_{\tilde{\texttt{proj}}}^t$ & 40.68 & 0.996 &	0.0117 \\
            \midrule
            w.o. MLP for skin. & 40.58 & 0.996	& 0.0112 \\
            w.o. detail def. & 39.02 &	0.993	 & 0.0163\\
            \midrule
            Anisotropy & 41.98 & 0.996 & 0.0080 \\
            Isotropy (Ours) & 40.82 & 0.996 & 0.0112 \\
            \bottomrule
		\end{tabular}
	\end{small}
\end{table}

%% file: figures_scripts/ablation_motion.tex
\begin{figure}[h] 
    \centering
\includegraphics[width=1\columnwidth]{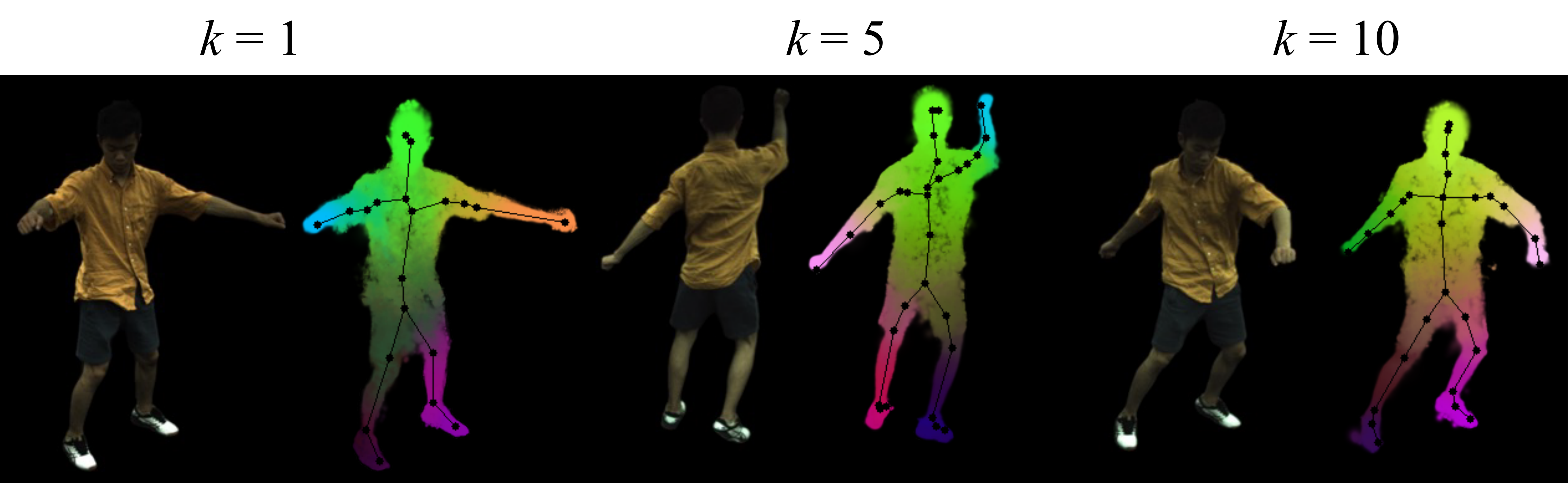} 
    \caption{The input image (left) and the visualized skeleton and skinning weights (right).}
    \label{fig:ablas_large_motion} 
\end{figure}

%% file: figures_scripts/supp_scgs_edit.tex
\begin{figure}[h] 
    \centering
\includegraphics[width=1.0\columnwidth]{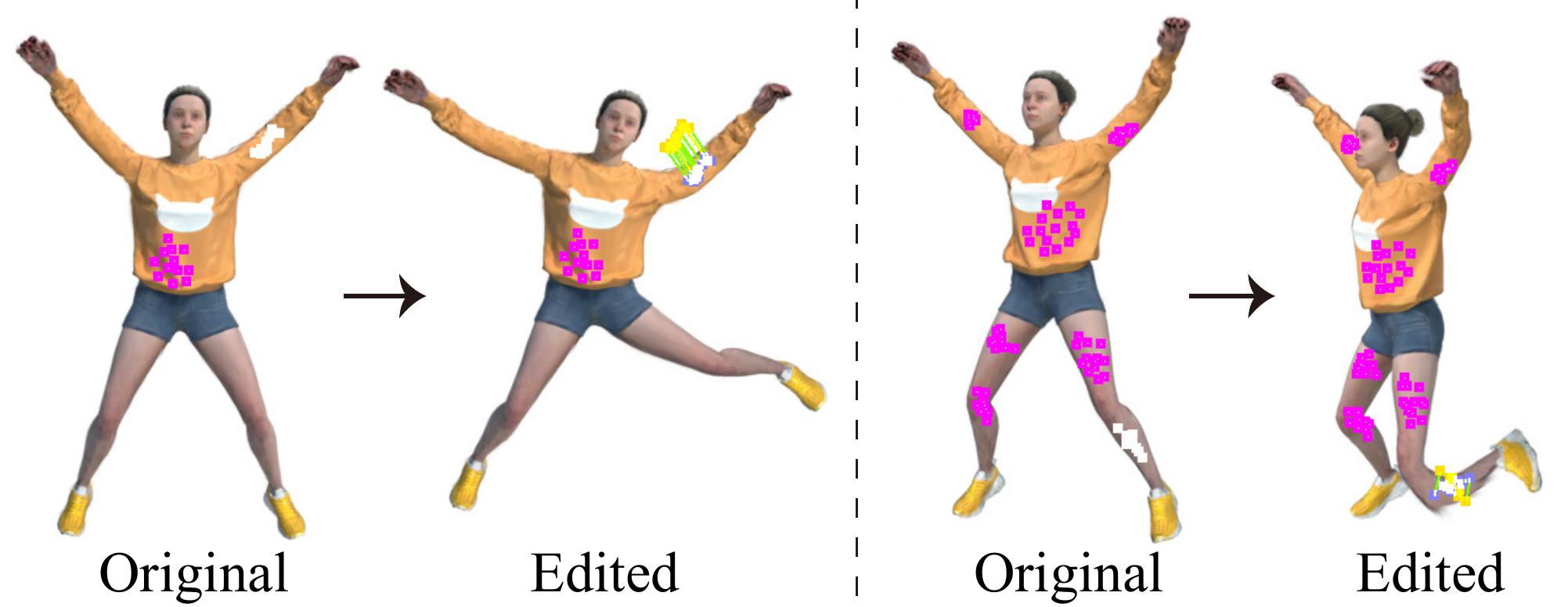}
    \caption{Editing by SC-GS~\cite{huang2024sc}. The pink dots mark the fixed locations, while the white dots indicate the positions to be edited. The green lines represent the trajectories of the edits.}
    \label{fig:supp_scgs_edit} 
\end{figure}

%% file: figures_scripts/supp_compare_zju.tex
\begin{figure*}[h] 
    \centering
\includegraphics[width=0.9\textwidth]{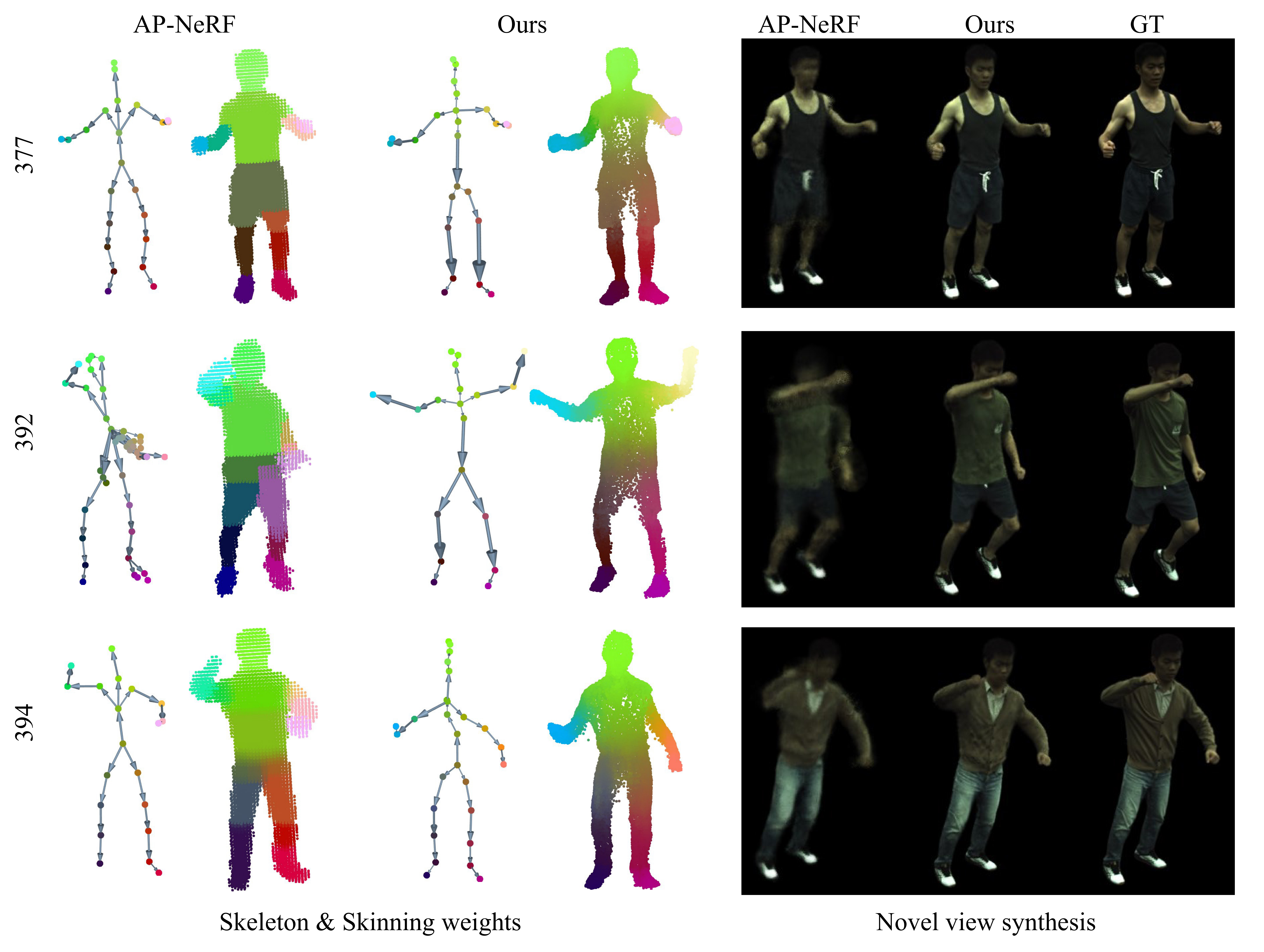} 
    \caption{Comparisons of skeleton, skinning weights and novel view synthesis with AP-NeRF on the ZJU-MoCap dataset~\cite{peng2021neural}.}
    \label{fig:compare_zju} 
\end{figure*}

%% file: table_scripts/supp_compare_zju.tex
\begin{table*}[h]
	\caption{Comparisons of the average precision on the ZJU-MoCap dataset~\cite{peng2021neural}.}
	\label{Tab:supp_compare_zju}
		\setlength{\tabcolsep}{3.8pt}
	\centering
	\begin{small}
		\begin{tabular}{ c   c  c c c c c c c c}
            \toprule 
            \multirow{2}{*}{\makecell[c]{Method}}  & \multicolumn{3}{c}{377} & \multicolumn{3}{c}{386} & \multicolumn{3}{c}{387} \\
            \cmidrule(r){2-4} \cmidrule(r){5-7} \cmidrule(r){8-10} 
             &  PSNR $\uparrow$ &  SSIM $\uparrow$ & LPIPS $\downarrow$ & PSNR $\uparrow$ &  SSIM $\uparrow$ & LPIPS $\downarrow$ &  PSNR $\uparrow$ &  SSIM $\uparrow$ & LPIPS $\downarrow$ \\  
            \midrule
            AP-NeRF~\cite{uzolas2024template} &  24.39 & 0.925 & 0.0827  & 28.94 & 0.932 & 0.0841  & 24.30 & 0.917 & 0.0961 \\
            Ours & 33.78 & 0.983 & 0.0202  & 36.63 & 0.981 & 0.0285  & 31.25 & 0.971 & 0.0395  \\\midrule 
            \multirow{2}{*}{\makecell[c]{Method}}  & \multicolumn{3}{c}{392} & \multicolumn{3}{c}{393} & \multicolumn{3}{c}{394} \\
            \cmidrule(r){2-4} \cmidrule(r){5-7} \cmidrule(r){8-10}  
            &  PSNR $\uparrow$ &  SSIM $\uparrow$ & LPIPS $\downarrow$ & PSNR $\uparrow$ &  SSIM $\uparrow$ & LPIPS $\downarrow$ &  PSNR $\uparrow$ &  SSIM $\uparrow$ & LPIPS $\downarrow$ \\\midrule  
            AP-NeRF~\cite{uzolas2024template} & 26.08 & 0.918 & 0.0983  & 24.53 & 0.908 & 0.1050  & 25.49 & 0.915 & 0.0943  \\
            Ours & 34.06 & 0.975 & 0.0351  & 31.65 & 0.969 & 0.0391  &  33.87 & 0.974 & 0.0339 \\
            \bottomrule
		\end{tabular}
	\end{small}
\end{table*}

%% file: table_scripts/supp_compare_blender.tex
\begin{table*}[h]
	\caption{Comparisons with the state-of-the-art methods on the D-NeRF dataset~\cite{pumarola2021dnerf}. The best and second-best results are highlighted in bold and underlined.}
	\label{Tab:comparisons-dnerf}
		\setlength{\tabcolsep}{3.8pt}
	\centering
	\begin{small}
		\begin{tabular}{ c c  c c c c c c c c c }
            \toprule 
            \multirow{2}{*}{\makecell[c]{Method}}  &  \multirow{2}{*}{\makecell[c]{Skeleton}} & \multicolumn{3}{c}{Hook} & \multicolumn{3}{c}{Trex} & \multicolumn{3}{c}{JumpingJacks} \\
            \cmidrule(r){3-5} \cmidrule(r){6-8} \cmidrule(r){9-11} 
            & & PSNR $\uparrow$  & SSIM $\uparrow$ & LPIPS $\downarrow$ & PSNR $\uparrow$  & SSIM $\uparrow$ & LPIPS $\downarrow$ & PSNR $\uparrow$  & SSIM $\uparrow$ & LPIPS $\downarrow$ \\
            \midrule
            D-NeRF~\cite{pumarola2021dnerf} & No & 29.25 & 0.968 & 0.1120 & 31.75 & 0.974 & 0.0367  & 32.80 & 0.981 & 0.0381\\
            TiNeuVox~\cite{fang2022TiNeuVox} & No & 31.45 & 0.971 & 0.0569 & 32.70 & 0.987 & 0.0340  & 34.23 & 0.986 & 0.0383 \\
            4D-GS~\cite{wu20244dgs} & No & 30.99 & 0.990 & 0.0248 & 32.16 & \underline{0.988} & 0.0216  & 33.59 & 0.990 & 0.0242 \\
            SC-GS~\cite{huang2024sc} & No & \textbf{39.87} & \textbf{0.997} & \textbf{0.0076}   & \textbf{41.24} & \textbf{0.998} & \textbf{0.0046} & \textbf{41.13} & \textbf{0.998} & \textbf{0.0067} \\
            \midrule 
            AP-NeRF~\cite{uzolas2024template} & Yes & 30.24 & 0.970 & 0.0500 & 32.85 & 0.980 & 0.0200 & 34.50 & 0.980 & 0.0300\\
            Ours & Yes & \underline{37.49} & \underline{0.994} & \underline{0.0136}   & \underline{38.40} & \textbf{0.998} & \underline{0.0063} & \underline{40.70} & \underline{0.997} & \underline{0.0069} \\
            \bottomrule
            \multirow{2}{*}{\makecell[c]{Method}}  &  \multirow{2}{*}{\makecell[c]{Skeleton}} & \multicolumn{3}{c}{Hellwarrior} & \multicolumn{3}{c}{Mutant} & \multicolumn{3}{c}{Standup} \\
            \cmidrule(r){3-5} \cmidrule(r){6-8} \cmidrule(r){9-11} 
            & & PSNR $\uparrow$  & SSIM $\uparrow$ & LPIPS $\downarrow$ & PSNR $\uparrow$  & SSIM $\uparrow$ & LPIPS $\downarrow$ & PSNR $\uparrow$  & SSIM $\uparrow$ & LPIPS $\downarrow$ \\
            \midrule
            D-NeRF~\cite{pumarola2021dnerf} & No & 25.02 & 0.955 & 0.0633 & 31.29 & 0.978 & 0.0212  & 32.79 & 0.983 & 0.0241\\
            TiNeuVox~\cite{fang2022TiNeuVox} & No & 28.17 & 0.978 & 0.0706 & 33.61 & 0.982 & 0.0388  & 35.43 & 0.991 & 0.0230 \\
            4D-GS~\cite{wu20244dgs} & No & 31.39 & 0.974 & 0.0436 & 35.98 & 0.996 & 0.0120  & 35.37 & 0.994 & 0.0136 \\
            SC-GS~\cite{huang2024sc} & No & \textbf{42.93} & \textbf{0.994} & \textbf{0.0155} & \textbf{45.19}  & \textbf{0.999} & \textbf{0.0028}  & \textbf{47.89} & \textbf{0.999} & \textbf{0.0023}\\
            \midrule 
            AP-NeRF~\cite{uzolas2024template} & Yes & 27.53 & 0.960 & 0.0600 & 28.56  &0.960 & 0.0300& 31.93 & 0.970 & 0.0200 \\
            Ours & Yes & \underline{41.21} & \underline{0.989} & \underline{0.0301} & \underline{42.72} & \underline{0.998} & \underline{0.0057} & \underline{44.37} & \underline{0.998} & \underline{0.0047} \\
            \bottomrule
		\end{tabular}
	\end{small}
\end{table*} 

%% file: figures_scripts/supp_failure_case.tex
\begin{figure}[h] 
    \centering
    \vspace{-0.1cm}
\includegraphics[width=0.9\columnwidth]{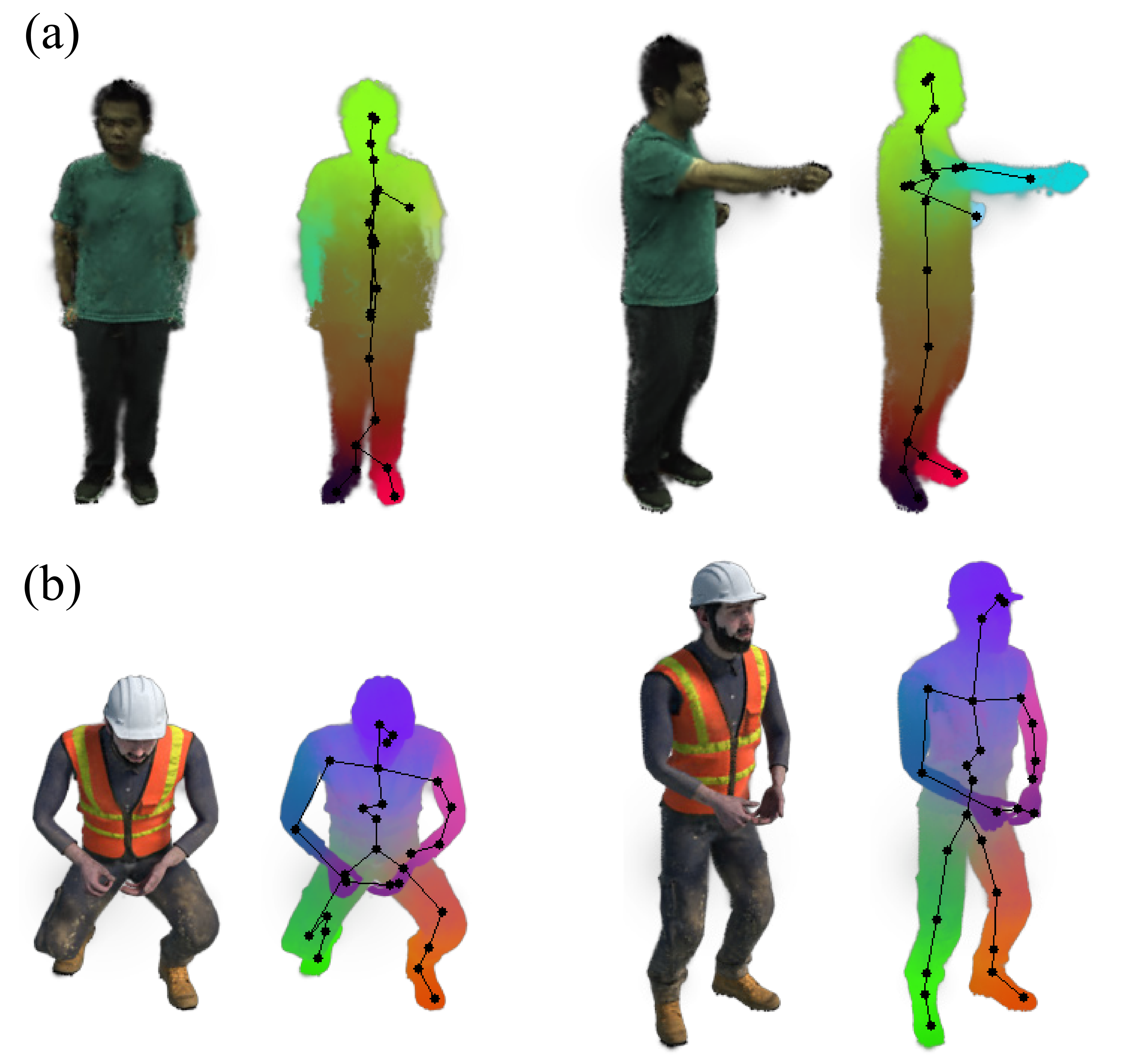}
    \caption{Two frames of rendered images (left), skeleton and skinning weights (right) from (a) ``386'' on the ZJU-MoCap dataset, and (b) ``Standup'' on the D-NeRF dataset.}
    \label{fig:supp_failure_case} 
\end{figure}

%% file: figures_scripts/supp_skeleton.tex
\begin{figure*}[h] 
    \centering
\includegraphics[width=\textwidth]{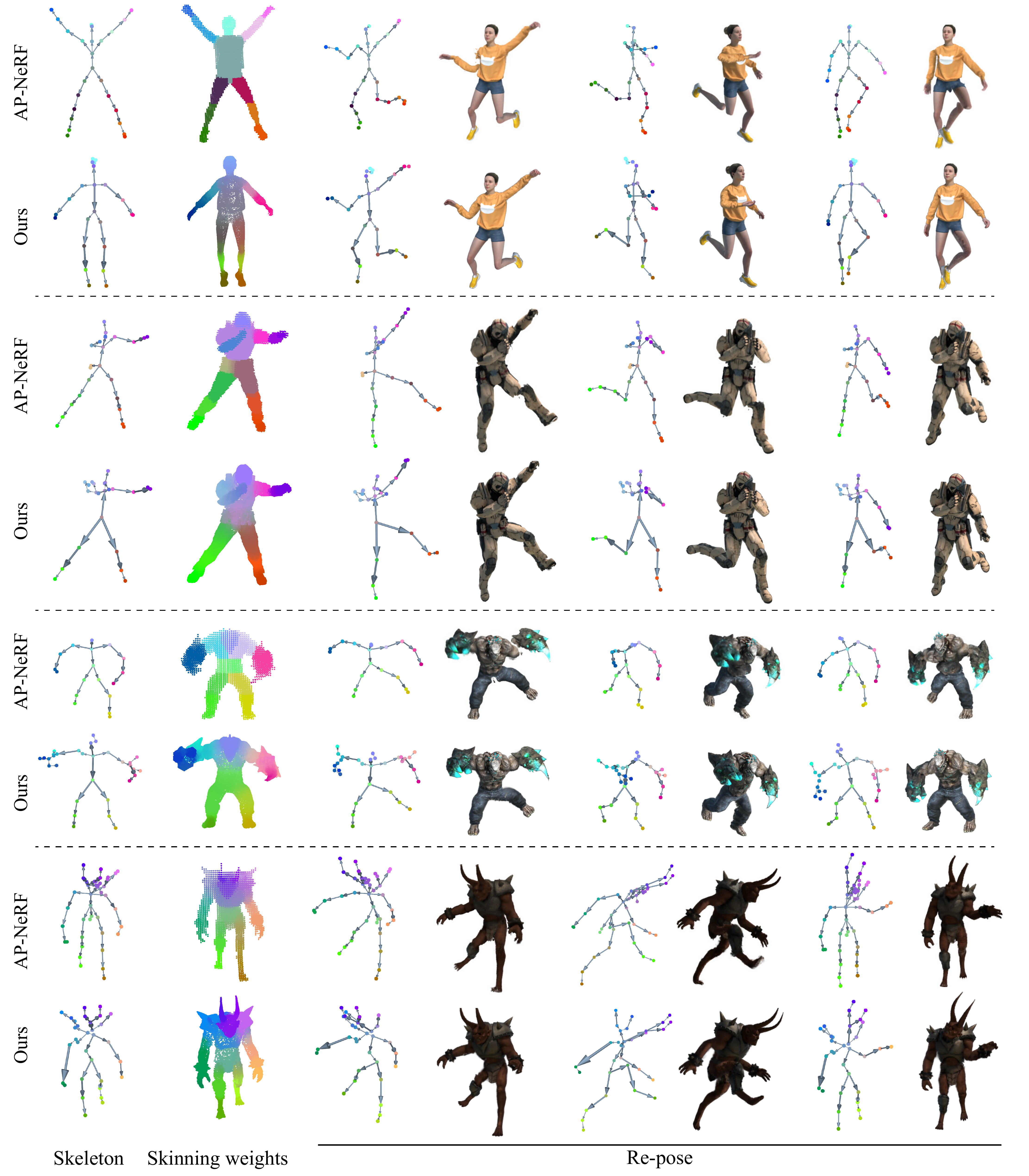}
    \caption{Editing on the D-NeRF dataset.}
    \label{fig:supp_skeleton} 
\end{figure*}

%% file: figures_scripts/supp_skeleton_dgmesh.tex
\begin{figure*}[h] 
    \centering
\includegraphics[width=\textwidth]{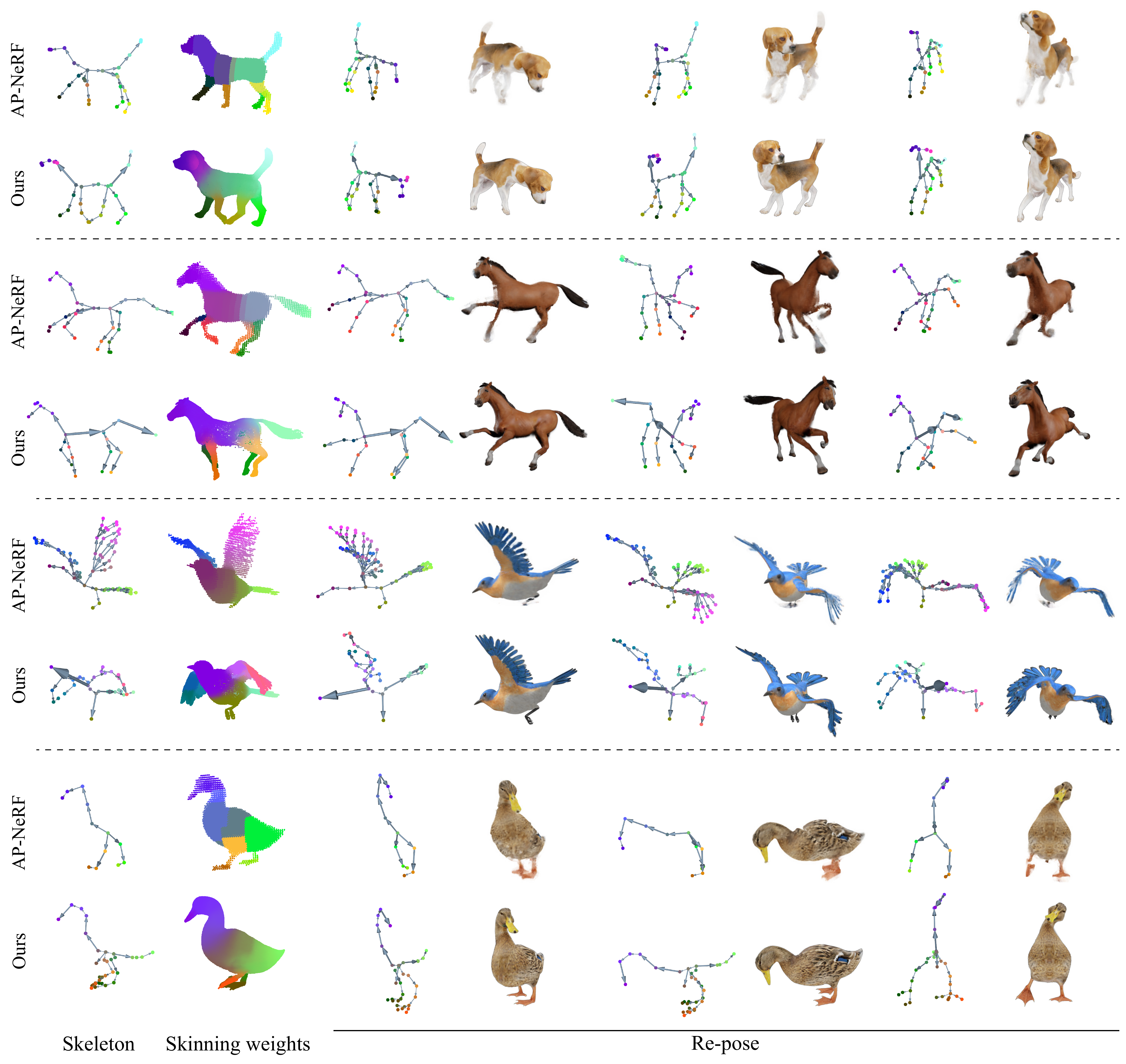}
    \caption{Editing on the DG-Mesh dataset.}
    \label{fig:supp_skeleton_dgmesh} 
\end{figure*}

%% file: figures_scripts/supp_nvs.tex
\begin{figure*}[h] 
    \centering
\includegraphics[width=\textwidth]{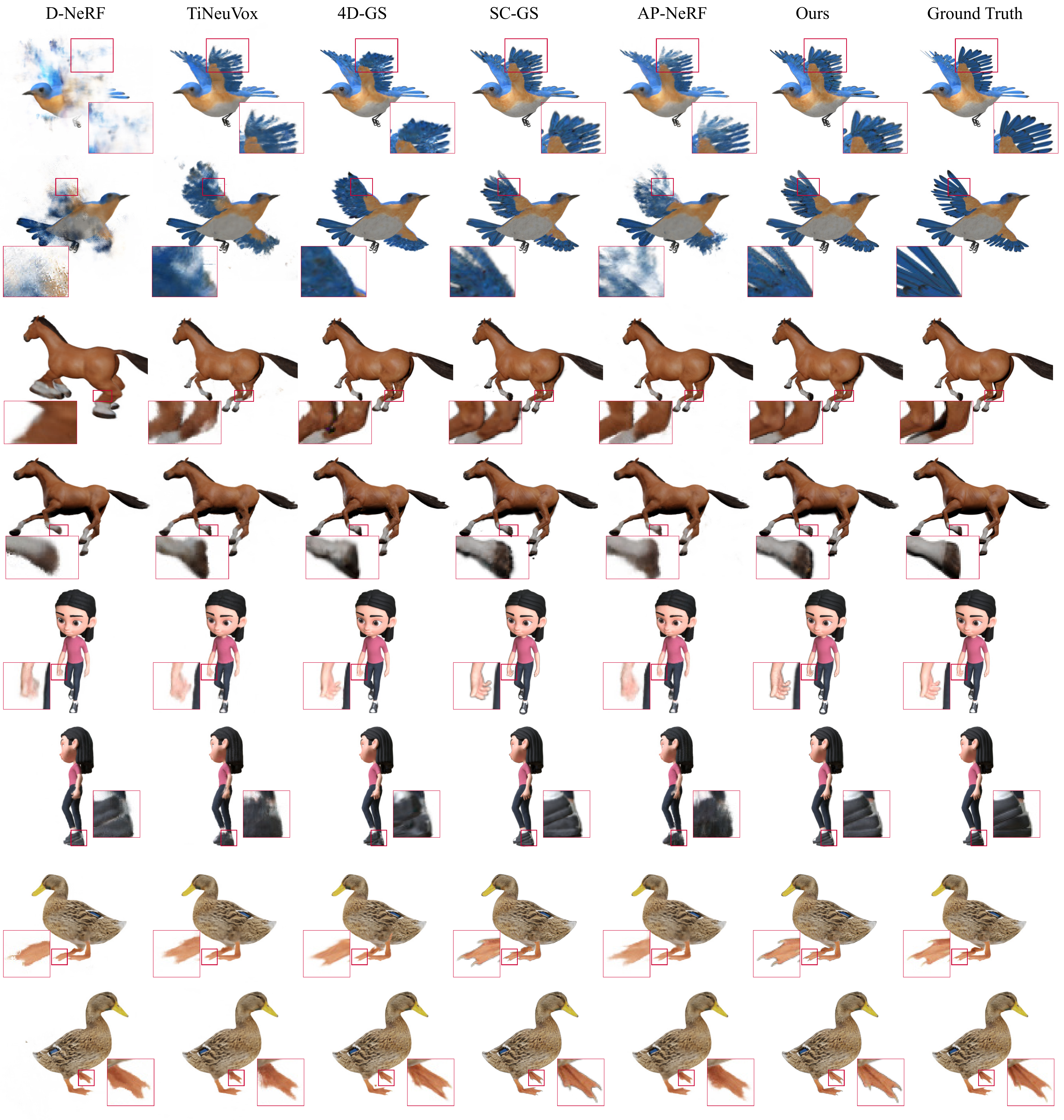}
    \caption{Novel view rendering on the DG-Mesh dataset.}
    \label{fig:supp_nvs} 
\end{figure*}

%% file: table_scripts/supp_compare_dgmesh.tex
\begin{table*}[h]
	\caption{Comparisons with the state-of-the-art methods on the DG-Mesh dataset~\cite{liu2024dynamic}. The best and second-best results are highlighted in bold and underlined.}
	\label{Tab:comparisons-dgmesh}
		\setlength{\tabcolsep}{3pt}
	\centering
	\begin{small}
		\begin{tabular}{ c c  c c c c c c c c c }
            \toprule 
            \multirow{2}{*}{\makecell[c]{Method}}  &  \multirow{2}{*}{\makecell[c]{Skeleton}} & \multicolumn{3}{c}{Beagle} & \multicolumn{3}{c}{Bird} & \multicolumn{3}{c}{Duck} \\
            \cmidrule(r){3-5} \cmidrule(r){6-8} \cmidrule(r){9-11} 
            & & PSNR $\uparrow$  & SSIM $\uparrow$ & LPIPS $\downarrow$ & PSNR $\uparrow$  & SSIM $\uparrow$ & LPIPS $\downarrow$ & PSNR $\uparrow$  & SSIM $\uparrow$ & LPIPS $\downarrow$ \\
            \midrule
            D-NeRF~\cite{pumarola2021dnerf} & No & -- & -- & -- & 21.05 & 0.884 & 0.1890  & 32.71 & 0.982 & 0.0312\\
            TiNeuVox~\cite{fang2022TiNeuVox} & No & 38.86 & 0.983 & 0.0287 & 25.69 & 0.934 & 0.0841  & 34.38 & 0.973 & 0.0291 \\
            4D-GS~\cite{wu20244dgs} & No & \textbf{42.15} & \underline{0.990} & 0.0222 & 26.75 & \underline{0.958} & 0.0443  & 36.69 & 0.984 & 0.0193 \\
            SC-GS~\cite{huang2024sc} & No & \underline{41.20} & \textbf{0.998} & \textbf{0.0054} & \textbf{32.55} & \textbf{0.980} & \underline{0.0269}  & \textbf{40.41} & \textbf{0.998} & \textbf{0.0047}\\
            \midrule 
            AP-NeRF~\cite{uzolas2024template} & Yes &  38.70 & 0.984  &  0.0281 &  25.08  &  0.933 & 0.0827  &  34.17 & 0.973  & 0.0287 \\
            Ours & Yes & 39.74 & \textbf{0.998} & \underline{0.0077}   & \underline{31.82} & \textbf{0.980} & \textbf{0.0263} & \underline{39.84} & \underline{0.997} & \underline{0.0061}\\
              \bottomrule
              \multirow{2}{*}{\makecell[c]{Method}}  &  \multirow{2}{*}{\makecell[c]{Skeleton}} & \multicolumn{3}{c}{Girlwalk} & \multicolumn{3}{c}{Horse} & \multicolumn{3}{c}{Average} \\
            \cmidrule(r){3-5} \cmidrule(r){6-8} \cmidrule(r){9-11} 
            & & PSNR $\uparrow$  & SSIM $\uparrow$ & LPIPS $\downarrow$ & PSNR $\uparrow$  & SSIM $\uparrow$ & LPIPS $\downarrow$ & PSNR $\uparrow$  & SSIM $\uparrow$ & LPIPS $\downarrow$ \\
            \midrule
            D-NeRF~\cite{pumarola2021dnerf} & No & 31.15 & 0.989 & 0.0336 & 27.78 & 0.971 & 0.0573  & 28.17 & 0.957 & 0.0778 \\
            TiNeuVox~\cite{fang2022TiNeuVox} & No & 32.62 & 0.984 & 0.0341 & 28.18 & 0.960 & 0.0623  & 31.95 & 0.967 & 0.0477 \\
            4D-GS~\cite{wu20244dgs} & No & 34.15 & 0.989 & 0.0145 & 30.07 & 0.974 & 0.0357  & 33.96 & 0.979 & 0.0272 \\
            SC-GS~\cite{huang2024sc} & No & \textbf{42.33} & \textbf{0.998} & \textbf{0.0084} & \textbf{38.29} & \textbf{0.990} & \textbf{0.0227}  & \textbf{38.96} & \textbf{0.993} & \textbf{0.0136}\\
            \midrule 
            AP-NeRF~\cite{uzolas2024template} & Yes &  32.63 &  0.984 & 0.0344  &  28.58 &  0.963 &  0.0561 & 31.83	&0.967 &	0.0460\\
            Ours & Yes & \underline{40.98} & \underline{0.997} & \underline{0.0107} & \underline{35.87} & \underline{0.984} & \underline{0.0337} & \underline{37.65} & \underline{0.991} & \underline{0.0169} \\
            \bottomrule
		\end{tabular}
	\end{small}
\end{table*}

%% file: main.bbl
\begin{thebibliography}{49}
\providecommand{\natexlab}[1]{#1}
\providecommand{\url}[1]{\texttt{#1}}
\expandafter\ifx\csname urlstyle\endcsname\relax
  \providecommand{\doi}[1]{doi: #1}\else
  \providecommand{\doi}{doi: \begingroup \urlstyle{rm}\Url}\fi

\bibitem[Bae et~al.(2022)Bae, Jang, Min, Choi, and Kim]{bae2022neural}
Jinseok Bae, Hojun Jang, Cheol-Hui Min, Hyungun Choi, and Young~Min Kim.
\newblock Neural marionette: Unsupervised learning of motion skeleton and latent dynamics from volumetric video.
\newblock In \emph{Proceedings of the AAAI Conference on Artificial Intelligence}, pages 86--94, 2022.

\bibitem[Bo{\v{z}}i{\v{c}} et~al.(2021)Bo{\v{z}}i{\v{c}}, Palafox, Zollh{\"o}fer, Thies, Dai, and Nie{\ss}ner]{bozic2021ndg}
Alja{\v{z}} Bo{\v{z}}i{\v{c}}, Pablo Palafox, Michael Zollh{\"o}fer, Justus Thies, Angela Dai, and Matthias Nie{\ss}ner.
\newblock Neural deformation graphs for globally-consistent non-rigid reconstruction.
\newblock In \emph{IEEE Conf. Comput. Vis. Pattern Recog.}, 2021.

\bibitem[Diederik(2014)]{diederik2014adam}
P~Kingma Diederik.
\newblock Adam: A method for stochastic optimization.
\newblock \emph{(No Title)}, 2014.

\bibitem[Dou et~al.(2022)Dou, Lin, Xu, Yang, Xin, Komura, and Wang]{dou2022coverage}
Zhiyang Dou, Cheng Lin, Rui Xu, Lei Yang, Shiqing Xin, Taku Komura, and Wenping Wang.
\newblock Coverage axis: Inner point selection for 3d shape skeletonization.
\newblock In \emph{Comput. Graph. Forum}, pages 419--432. Wiley Online Library, 2022.

\bibitem[Fan et~al.(2017)Fan, Su, and Guibas]{fan2017point}
Haoqiang Fan, Hao Su, and Leonidas~J Guibas.
\newblock A point set generation network for 3d object reconstruction from a single image.
\newblock In \emph{IEEE Conf. Comput. Vis. Pattern Recog.}, pages 605--613, 2017.

\bibitem[Fang et~al.(2022)Fang, Yi, Wang, Xie, Zhang, Liu, Nie\ss{}ner, and Tian]{fang2022TiNeuVox}
Jiemin Fang, Taoran Yi, Xinggang Wang, Lingxi Xie, Xiaopeng Zhang, Wenyu Liu, Matthias Nie\ss{}ner, and Qi Tian.
\newblock Fast dynamic radiance fields with time-aware neural voxels.
\newblock In \emph{SIGGRAPH Asia 2022 Conference Papers}, 2022.

\bibitem[Huang et~al.(2024)Huang, Sun, Yang, Lyu, Cao, and Qi]{huang2024sc}
Yi-Hua Huang, Yang-Tian Sun, Ziyi Yang, Xiaoyang Lyu, Yan-Pei Cao, and Xiaojuan Qi.
\newblock Sc-gs: Sparse-controlled gaussian splatting for editable dynamic scenes.
\newblock In \emph{IEEE Conf. Comput. Vis. Pattern Recog.}, pages 4220--4230, 2024.

\bibitem[Jiang et~al.(2022)Jiang, Hong, Bao, and Zhang]{jiang2022selfrecon}
Boyi Jiang, Yang Hong, Hujun Bao, and Juyong Zhang.
\newblock Selfrecon: Self reconstruction your digital avatar from monocular video.
\newblock In \emph{IEEE Conf. Comput. Vis. Pattern Recog.}, 2022.

\bibitem[Kerbl et~al.(2023{\natexlab{a}})Kerbl, Kopanas, Leimk{\"u}hler, and Drettakis]{kerbl20233d}
Bernhard Kerbl, Georgios Kopanas, Thomas Leimk{\"u}hler, and George Drettakis.
\newblock 3d gaussian splatting for real-time radiance field rendering.
\newblock \emph{ACM Trans. Graph.}, 42\penalty0 (4):\penalty0 139--1, 2023{\natexlab{a}}.

\bibitem[Kerbl et~al.(2023{\natexlab{b}})Kerbl, Kopanas, Leimk{\"u}hler, and Drettakis]{kerbl20233dgs}
Bernhard Kerbl, Georgios Kopanas, Thomas Leimk{\"u}hler, and George Drettakis.
\newblock 3d gaussian splatting for real-time radiance field rendering.
\newblock \emph{ACM Trans. Graph.}, 42\penalty0 (4):\penalty0 139--1, 2023{\natexlab{b}}.

\bibitem[Kuai et~al.(2023)Kuai, Karthikeyan, Kant, Mirzaei, and Gilitschenski]{kuai2023camm}
Tianshu Kuai, Akash Karthikeyan, Yash Kant, Ashkan Mirzaei, and Igor Gilitschenski.
\newblock Camm: Building category-agnostic and animatable 3d models from monocular videos.
\newblock In \emph{Proceedings of the IEEE/CVF Conference on Computer Vision and Pattern Recognition Workshops}, pages 6586--6596, 2023.

\bibitem[Lewis et~al.(2000)Lewis, Cordner, and Fong]{lewis2023pose}
J.~P. Lewis, Matt Cordner, and Nickson Fong.
\newblock Pose space deformation: a unified approach to shape interpolation and skeleton-driven deformation.
\newblock In \emph{Proceedings of the 27th Annual Conference on Computer Graphics and Interactive Techniques}, page 165–172, USA, 2000. ACM Press/Addison-Wesley Publishing Co.

\bibitem[Lin et~al.(2021)Lin, Li, Liu, Chen, Choi, and Wang]{lin2021point2skeleton}
Cheng Lin, Changjian Li, Yuan Liu, Nenglun Chen, Yi-King Choi, and Wenping Wang.
\newblock Point2skeleton: Learning skeletal representations from point clouds.
\newblock In \emph{IEEE Conf. Comput. Vis. Pattern Recog.}, pages 4277--4286, 2021.

\bibitem[Liu et~al.(2025{\natexlab{a}})Liu, Su, and Wang]{liu2024dynamic}
Isabella Liu, Hao Su, and Xiaolong Wang.
\newblock Dynamic gaussians mesh: Consistent mesh reconstruction from monocular videos.
\newblock In \emph{Int. Conf. Learn. Represent.}, 2025{\natexlab{a}}.

\bibitem[Liu et~al.(2025{\natexlab{b}})Liu, Liu, Wang, Lyv, Wang, Wang, and Hou]{liu2024modgs}
Qingming Liu, Yuan Liu, Jiepeng Wang, Xianqiang Lyv, Peng Wang, Wenping Wang, and Junhui Hou.
\newblock Modgs: Dynamic gaussian splatting from casually-captured monocular videos.
\newblock In \emph{Int. Conf. Learn. Represent.}, 2025{\natexlab{b}}.

\bibitem[Liu et~al.(2023)Liu, Gupta, and Wang]{liu2023reart}
Shaowei Liu, Saurabh Gupta, and Shenlong Wang.
\newblock Building rearticulable models for arbitrary 3d objects from 4d point clouds.
\newblock In \emph{IEEE Conf. Comput. Vis. Pattern Recog.}, 2023.

\bibitem[Loper et~al.(2015)Loper, Mahmood, Romero, Pons-Moll, and Black]{loper15smpl}
Matthew Loper, Naureen Mahmood, Javier Romero, Gerard Pons-Moll, and Michael~J. Black.
\newblock Smpl: a skinned multi-person linear model.
\newblock \emph{ACM Trans. Graph.}, 34\penalty0 (6), 2015.

\bibitem[Mildenhall et~al.(2020)Mildenhall, Srinivasan, Tancik, Barron, Ramamoorthi, and Ng]{mildenhall2020nerf}
Ben Mildenhall, Pratul~P. Srinivasan, Matthew Tancik, Jonathan~T. Barron, Ravi Ramamoorthi, and Ren Ng.
\newblock Nerf: Representing scenes as neural radiance fields for view synthesis.
\newblock In \emph{Eur. Conf. Comput. Vis.}, 2020.

\bibitem[Models(2024)]{lei2024gart}
{GART:} Gaussian Articulated~Template Models.
\newblock Jiahui lei and yufu wang and georgios pavlakos and lingjie liu and kostas daniilidis.
\newblock In \emph{IEEE Conf. Comput. Vis. Pattern Recog.}, 2024.

\bibitem[Moon et~al.(2024)Moon, Shiratori, and Saito]{moon2024exavatar}
Gyeongsik Moon, Takaaki Shiratori, and Shunsuke Saito.
\newblock Expressive whole-body {3D} gaussian avatar.
\newblock In \emph{Eur. Conf. Comput. Vis.}, 2024.

\bibitem[Oquab et~al.(2023)Oquab, Darcet, Moutakanni, Vo, Szafraniec, Khalidov, Fernandez, Haziza, Massa, El-Nouby, Howes, Huang, Xu, Sharma, Li, Galuba, Rabbat, Assran, Ballas, Synnaeve, Misra, Jegou, Mairal, Labatut, Joulin, and Bojanowski]{oquab2023dinov2}
Maxime Oquab, Timothée Darcet, Theo Moutakanni, Huy~V. Vo, Marc Szafraniec, Vasil Khalidov, Pierre Fernandez, Daniel Haziza, Francisco Massa, Alaaeldin El-Nouby, Russell Howes, Po-Yao Huang, Hu Xu, Vasu Sharma, Shang-Wen Li, Wojciech Galuba, Mike Rabbat, Mido Assran, Nicolas Ballas, Gabriel Synnaeve, Ishan Misra, Herve Jegou, Julien Mairal, Patrick Labatut, Armand Joulin, and Piotr Bojanowski.
\newblock Dinov2: Learning robust visual features without supervision, 2023.

\bibitem[Peng et~al.(2021)Peng, Zhang, Xu, Wang, Shuai, Bao, and Zhou]{peng2021neural}
Sida Peng, Yuanqing Zhang, Yinghao Xu, Qianqian Wang, Qing Shuai, Hujun Bao, and Xiaowei Zhou.
\newblock Neural body: Implicit neural representations with structured latent codes for novel view synthesis of dynamic humans.
\newblock In \emph{IEEE Conf. Comput. Vis. Pattern Recog.}, 2021.

\bibitem[Pumarola et~al.(2021)Pumarola, Corona, Pons-Moll, and Moreno-Noguer]{pumarola2021dnerf}
Albert Pumarola, Enric Corona, Gerard Pons-Moll, and Francesc Moreno-Noguer.
\newblock D-nerf: Neural radiance fields for dynamic scenes.
\newblock In \emph{IEEE Conf. Comput. Vis. Pattern Recog.}, pages 10318--10327, 2021.

\bibitem[Romero et~al.(2017)Romero, Tzionas, and Black]{romero2017mano}
Javier Romero, Dimitrios Tzionas, and Michael~J. Black.
\newblock Embodied hands: Modeling and capturing hands and bodies together.
\newblock \emph{ACM Trans. Graph.}, 36\penalty0 (6), 2017.

\bibitem[Song et~al.(2024)Song, Wei, Foo, Lin, and Liu]{song2024reacto}
Chaoyue Song, Jiacheng Wei, Chuan~Sheng Foo, Guosheng Lin, and Fayao Liu.
\newblock Reacto: Reconstructing articulated objects from a single video.
\newblock In \emph{IEEE Conf. Comput. Vis. Pattern Recog.}, pages 5384--5395, 2024.

\bibitem[Sorkine and Alexa(2007)]{sorkine2007rigid}
Olga Sorkine and Marc Alexa.
\newblock As-rigid-as-possible surface modeling.
\newblock In \emph{Symposium on Geometry processing}, pages 109--116. Citeseer, 2007.

\bibitem[Tu et~al.(2023)Tu, Li, Lin, Cheng, Sun, and Yang]{tu2023dreamo}
Tao Tu, Ming-Feng Li, Chieh~Hubert Lin, Yen-Chi Cheng, Min Sun, and Ming-Hsuan Yang.
\newblock Dreamo: Articulated 3d reconstruction from a single casual video.
\newblock \emph{arXiv preprint arXiv:2312.02617}, 2023.

\bibitem[Uzolas et~al.(2024)Uzolas, Eisemann, and Kellnhofer]{uzolas2024template}
Lukas Uzolas, Elmar Eisemann, and Petr Kellnhofer.
\newblock Template-free articulated neural point clouds for reposable view synthesis.
\newblock \emph{Adv. Neural Inform. Process. Syst.}, 36, 2024.

\bibitem[Wan et~al.(2024)Wan, Wang, Lu, and Zeng]{wan2024skgs}
Diwen Wan, Yuxiang Wang, Ruijie Lu, and Gang Zeng.
\newblock Template-free articulated gaussian splatting for real-time reposable dynamic view synthesis.
\newblock In \emph{Adv. Neural Inform. Process. Syst.}, 2024.

\bibitem[Wang et~al.(2021)Wang, Liu, Liu, Theobalt, Komura, and Wang]{wang2021neus}
Peng Wang, Lingjie Liu, Yuan Liu, Christian Theobalt, Taku Komura, and Wenping Wang.
\newblock Neus: Learning neural implicit surfaces by volume rendering for multi-view reconstruction.
\newblock \emph{Adv. Neural Inform. Process. Syst.}, 2021.

\bibitem[Wang et~al.(2024)Wang, Dou, Xu, Lin, Liu, Long, Xin, Komura, Yuan, and Wang]{wang2024coverage}
Zimeng Wang, Zhiyang Dou, Rui Xu, Cheng Lin, Yuan Liu, Xiaoxiao Long, Shiqing Xin, Taku Komura, Xiaoming Yuan, and Wenping Wang.
\newblock Coverage axis++: Efficient inner point selection for 3d shape skeletonization.
\newblock In \emph{Comput. Graph. Forum}, page e15143. Wiley Online Library, 2024.

\bibitem[Wu et~al.(2024)Wu, Yi, Fang, Xie, Zhang, Wei, Liu, Tian, and Wang]{wu20244dgs}
Guanjun Wu, Taoran Yi, Jiemin Fang, Lingxi Xie, Xiaopeng Zhang, Wei Wei, Wenyu Liu, Qi Tian, and Xinggang Wang.
\newblock 4d gaussian splatting for real-time dynamic scene rendering.
\newblock In \emph{IEEE Conf. Comput. Vis. Pattern Recog.}, pages 20310--20320, 2024.

\bibitem[Wu et~al.(2023)Wu, Li, Jakab, Rupprecht, and Vedaldi]{wu2023magicpony}
Shangzhe Wu, Ruining Li, Tomas Jakab, Christian Rupprecht, and Andrea Vedaldi.
\newblock Magicpony: Learning articulated 3d animals in the wild.
\newblock In \emph{IEEE Conf. Comput. Vis. Pattern Recog.}, pages 8792--8802, 2023.

\bibitem[Wu* et~al.(2022)Wu*, Chen*, Liu, Ren, and Wang]{wu2022casa}
Yuefan Wu*, Zeyuan Chen*, Shaowei Liu, Zhongzheng Ren, and Shenlong Wang.
\newblock {CASA}: Category-agnostic skeletal animal reconstruction.
\newblock In \emph{Adv. Neural Inform. Process. Syst.}, 2022.

\bibitem[Xu et~al.(2019)Xu, Zhou, Kalogerakis, and Singh]{xu2019predicting}
Zhan Xu, Yang Zhou, Evangelos Kalogerakis, and Karan Singh.
\newblock Predicting animation skeletons for 3d articulated models via volumetric nets.
\newblock In \emph{2019 international conference on 3D vision (3DV)}, pages 298--307. IEEE, 2019.

\bibitem[Xu et~al.(2020)Xu, Zhou, Kalogerakis, Landreth, and Singh]{xu2020rignet}
Zhan Xu, Yang Zhou, Evangelos Kalogerakis, Chris Landreth, and Karan Singh.
\newblock Rignet: Neural rigging for articulated characters.
\newblock \emph{ACM Trans. Graph.}, 39\penalty0 (4), 2020.

\bibitem[Xu et~al.(2022)Xu, Zhou, Yi, and Kalogerakis]{xu2022morig}
Zhan Xu, Yang Zhou, Li Yi, and Evangelos Kalogerakis.
\newblock Morig: Motion-aware rigging of character meshes from point clouds.
\newblock In \emph{SIGGRAPH Asia 2022 conference papers}, pages 1--9, 2022.

\bibitem[Yang et~al.(2022)Yang, Vo, Neverova, Ramanan, Vedaldi, and Joo]{yang2022banmo}
Gengshan Yang, Minh Vo, Natalia Neverova, Deva Ramanan, Andrea Vedaldi, and Hanbyul Joo.
\newblock Banmo: Building animatable 3d neural models from many casual videos.
\newblock In \emph{IEEE Conf. Comput. Vis. Pattern Recog.}, 2022.

\bibitem[Yang et~al.(2023{\natexlab{a}})Yang, Wang, Reddy, and Ramanan]{yang2023rac}
Gengshan Yang, Chaoyang Wang, N.~Dinesh Reddy, and Deva Ramanan.
\newblock Reconstructing animatable categories from videos.
\newblock In \emph{IEEE Conf. Comput. Vis. Pattern Recog.}, 2023{\natexlab{a}}.

\bibitem[Yang et~al.(2023{\natexlab{b}})Yang, Wang, Reddy, and Ramanan]{yang2023reconstructing}
Gengshan Yang, Chaoyang Wang, N~Dinesh Reddy, and Deva Ramanan.
\newblock Reconstructing animatable categories from videos.
\newblock In \emph{IEEE Conf. Comput. Vis. Pattern Recog.}, pages 16995--17005, 2023{\natexlab{b}}.

\bibitem[Yang et~al.(2024{\natexlab{a}})Yang, Gao, Zhou, Jiao, Zhang, and Jin]{yang2024deformable}
Ziyi Yang, Xinyu Gao, Wen Zhou, Shaohui Jiao, Yuqing Zhang, and Xiaogang Jin.
\newblock Deformable 3d gaussians for high-fidelity monocular dynamic scene reconstruction.
\newblock In \emph{IEEE Conf. Comput. Vis. Pattern Recog.}, pages 20331--20341, 2024{\natexlab{a}}.

\bibitem[Yang et~al.(2024{\natexlab{b}})Yang, Zhou, Shan, Wen, Xuan, Hill, Bai, Qi, and Wang]{yang2024omnimotiongpt}
Zhangsihao Yang, Mingyuan Zhou, Mengyi Shan, Bingbing Wen, Ziwei Xuan, Mitch Hill, Junjie Bai, Guo-Jun Qi, and Yalin Wang.
\newblock Omnimotiongpt: Animal motion generation with limited data, 2024{\natexlab{b}}.

\bibitem[Yao et~al.(2022)Yao, Hung, Li, Rubinstein, Yang, and Jampani]{yao2022lassie}
Chun{-}Han Yao, Wei{-}Chih Hung, Yuanzhen Li, Michael Rubinstein, Ming{-}Hsuan Yang, and Varun Jampani.
\newblock {LASSIE:} learning articulated shapes from sparse image ensemble via 3d part discovery.
\newblock In \emph{Adv. Neural Inform. Process. Syst.}, 2022.

\bibitem[Yao et~al.(2024)Yao, Ren, Hou, Deng, Zhang, and Wang]{yao2024dynosurf}
Yuxin Yao, Siyu Ren, Junhui Hou, Zhi Deng, Juyong Zhang, and Wenping Wang.
\newblock Dynosurf: Neural deformation-based temporally consistent dynamic surface reconstruction.
\newblock In \emph{Eur. Conf. Comput. Vis.}, 2024.

\bibitem[You and Hou(2024)]{you2024decoupling}
Meng You and Junhui Hou.
\newblock Decoupling dynamic monocular videos for dynamic view synthesis.
\newblock \emph{IEEE Trans. Vis. Comput. Graph.}, 2024.

\bibitem[You et~al.(2025)You, Zhu, Liu, and Hou]{you2024nvs}
Meng You, Zhiyu Zhu, Hui Liu, and Junhui Hou.
\newblock Nvs-solver: Video diffusion model as zero-shot novel view synthesizer.
\newblock In \emph{Int. Conf. Learn. Represent.}, 2025.

\bibitem[Zhang et~al.(2024)Zhang, Gao, Li, Liu, and Chen]{zhang2024bags}
Tingyang Zhang, Qingzhe Gao, Weiyu Li, Libin Liu, and Baoquan Chen.
\newblock Bags: Building animatable gaussian splatting from a monocular video with diffusion priors, 2024.

\bibitem[Zhang and Suen(1984)]{zhang1984thinning}
T.~Y. Zhang and Ching~Y. Suen.
\newblock A fast parallel algorithm for thinning digital patterns.
\newblock \emph{Commun. {ACM}}, 27\penalty0 (3):\penalty0 236--239, 1984.

\bibitem[Zuffi et~al.(2017)Zuffi, Kanazawa, Jacobs, and Black]{zuffi2017smal}
Silvia Zuffi, Angjoo Kanazawa, David Jacobs, and Michael~J. Black.
\newblock {3D} menagerie: Modeling the {3D} shape and pose of animals.
\newblock In \emph{IEEE Conf. Comput. Vis. Pattern Recog.}, 2017.

\end{thebibliography}
